\documentclass{article}
\usepackage{iclr2027_conference,times}
\iclrfinalcopy
\usepackage[T1]{fontenc}
\usepackage[utf8]{inputenc}
\usepackage{amsmath,amssymb,amsthm,mathtools,bm}
\usepackage{graphicx,booktabs,array,tabularx,multirow}
\usepackage{xcolor}
\usepackage{microtype}
\usepackage{placeins}
\usepackage{float}
\usepackage{tikz}
\usetikzlibrary{arrows.meta,positioning,calc,angles,quotes}
\usepackage{hyperref}
\definecolor{linkblue}{rgb}{0.10,0.30,0.60}
\hypersetup{colorlinks=true,linkcolor=linkblue,citecolor=linkblue,urlcolor=linkblue,
 pdftitle={Learning Beyond Full Imitation: Task-Preserving Knowledge Distillation},
 pdfauthor={Qianfeng Yuan; Wenbing Tao},
 pdfkeywords={knowledge distillation, task preservation, conditional knowledge transfer, constrained optimization, learning dynamics}}
\usepackage{url}
\newcommand{\R}{\mathbb R}

\newcommand{\one}{\mathbf1}
\newcommand{\KL}{\operatorname{KL}}
\newcommand{\TV}{\operatorname{TV}}
\newcommand{\CE}{L_{\mathrm{CE}}}
\newcommand{\Shape}{L_{\mathrm{cond}}}
\newcommand{\Ber}{\operatorname{kl}_{\mathrm{Ber}}}
\newcommand{\softmax}{\operatorname{softmax}}
\newcommand{\logit}{\operatorname{logit}}

\newcommand{\Kc}{\mathcal K_y}
\newcommand{\Ls}{\mathcal L_y}
\newcommand{\norm}[1]{\left\lVert#1\right\rVert}
\newcommand{\ip}[2]{\left\langle#1,#2\right\rangle}
\newcommand{\stopgrad}{\operatorname{stopgrad}}

\newcolumntype{Y}{>{\raggedright\arraybackslash}X}

\newcommand{\tpkd}{TPKD}

\newtheorem{theorem}{Theorem}
\newtheorem{proposition}{Proposition}

\theoremstyle{remark}

\title{Learning Beyond Full Imitation:\\Task-Preserving Knowledge Distillation}
\author{Qianfeng Yuan \qquad Wenbing Tao\\
Huazhong University of Science and Technology}
\begin{document}
\maketitle
\lhead{Preprint}
\begin{abstract}
Knowledge distillation transfers knowledge by encouraging a student to match a teacher's predicted class probabilities. These probabilities express not only confidence in the correct class, but also relations among incorrect alternatives. Yet closer imitation does not necessarily yield a better student. Our key observation is that a student may already distinguish the correct class more sharply than its teacher, so further imitation can require giving back discrimination it has acquired.
Our main result is an exact separation between full imitation and conditional learning. When the correct class's score advantage over each alternative must be preserved, full teacher-to-student KL minimization is blocked exactly when the student assigns no more probability than the teacher to every incorrect class. Crucially, the teacher's relative probabilities among incorrect classes remain fully learnable. We characterize the exact price of this transfer: a minimum increase in correct-class log-odds that compensates for the largest conditional-probability mismatch. Label fitting and conditional matching can therefore be completed even as full teacher KL diverges.
This separation motivates task-preserving knowledge distillation (TPKD), which keeps the label gradient intact and minimally corrects the conditional gradient so that its output update preserves the label step's gains against every incorrect alternative. The corrected conditional direction retains more than half of the original first-order conditional descent at the same step size, with a tight bound. For a fixed positive conditional target and sufficiently small constant output steps, label and conditional errors vanish together. Experiments trace this learning from exact head updates to ordinary network training. TPKD reaches 88.05\% accuracy on CIFAR-100 and 93.81\% on CLINC150, improving over standard distillation by 0.47 and 0.35 percentage points across three seeds.

\end{abstract}
\noindent\textbf{Keywords:} knowledge distillation, task preservation, conditional knowledge transfer, constrained optimization, learning dynamics.

\section{Introduction}\label{sec:introduction}
Knowledge distillation (KD) transfers knowledge from a teacher to a student through the teacher's predictive distribution \citep{hinton2015distilling}. Besides confidence in the correct class, this distribution expresses how plausible the incorrect alternatives are relative to one another. These relative probabilities can encode similarities and confusions absent from hard labels, making them an important source of transferable knowledge \citep{zhao2022decoupled}.

Progress in KD has been driven largely by empirical advances in objectives and training procedures, while theoretical analyses remain comparatively limited \citep{phuong2019understanding,menon2021statistical,harutyunyan2023supervision}. A closer imitator is not necessarily a better student \citep{cho2019efficacy,stanton2021does}. Existing theory studies capacity, teacher quality, regularization, and data geometry or optimization bias. We study a complementary question: when further imitation conflicts with the student's task progress, what can the teacher still teach?

\textbf{Our key observation is that the student is not merely a passive copy of the teacher.} Label learning may make it more discriminative on an example, assigning lower probability than the teacher to every incorrect class. Moving its full prediction closer to the teacher then requires giving back some of that discrimination. Yet the teacher may still describe a different relative ordering among the incorrect alternatives. Reproducing the teacher's complete prediction and learning the structure within it are therefore different objectives.

This leads to the central question of this work: \emph{What can a student still learn from its teacher without sacrificing any of its existing task discrimination, and at what cost?}

Preserving all existing task discrimination requires more than keeping the answer correct. For example, a smaller score lead can make a previously accepted prediction require human confirmation \citep{liang2024selective}. Since different confusions may require different acceptance thresholds, we preserve each correct-versus-incorrect score gap---its margin---while allowing the teacher to reshape relations among incorrect classes.

\textbf{Our answer is an exact separation between full imitation and conditional learning.} We prove that full imitation is blocked precisely when the student assigns no more probability than the teacher to every incorrect class. Its current prediction is then already the closest point to the teacher within the entire task-preserving region. Crucially, this does not mean that the teacher has nothing left to teach. The student can still recover the teacher's entire conditional distribution over incorrect classes, not merely their ranking, without decreasing any correct-class margin.

Learning this structure has an exact compensation cost. Matching a different conditional distribution raises some incorrect alternatives relative to others. Preserving their margins requires a corresponding increase in correct-class confidence. \emph{The price is greater confidence, not weaker discrimination.} We derive the minimum increase in correct-class log-odds needed for exact, task-preserving transfer: the largest teacher-to-student conditional-probability ratio determines the cost, and a continuous safe path attains it. This yields an important consequence: the student can approach perfect label fitting while exactly matching the teacher's conditional structure, even as full teacher KL diverges. Learning the teacher's knowledge need not mean reproducing its confidence.

\textbf{This separation suggests how to keep learning.} We propose Task-Preserving Knowledge Distillation (TPKD). \textbf{Our main contributions are:} \textbf{Separation and transfer cost.} To our knowledge, we provide the first joint characterization of exact full-KL blocking, complete conditional reachability, and minimum confidence compensation under per-competitor margin preservation. We prove a necessary-and-sufficient condition for blocking and derive the attainable minimum correct-class log-odds cost. \textbf{A protected update with tight retention.} TPKD keeps the label gradient intact and adds the nearest safe conditional direction. It preserves every margin gain of the label step from the same start and with the same step size. The correction retains over half the first-order conditional descent with a tight bound, and a nonzero, aligned conditional component whenever error remains. \textbf{Joint learning and parameter-space guarantees.} For fixed positive targets and sufficiently small constant output steps, label and conditional errors vanish together. For nonuniform targets, this differs from CE's uniform conditional limit. Exact fixed-feature head updates and computable backpropagation conditions connect the output construction to network parameters.

Experiments test the mechanism and its use in full training. Across the evaluated visual CE states, blocking rises from 50.00\% to 58.90\%, with conditional knowledge still unlearned. All 64 exact-head batches gain additional conditional knowledge while retaining label-step margins; native updates carry this gain into the network and accumulate it over successive iterations. Full training reaches 88.05\% on CIFAR-100 and 93.81\% on CLINC150, exceeding standard KD by 0.47 and 0.35 percentage points across three seeds. Removing the CE gradient or replacing conditional learning with protected full imitation reduces accuracy in both domains, while removing projection gives closely matched performance.

\section{Related work}\label{sec:related}
\noindent\textbf{Distillation targets.} Classical KD matches softened teacher probabilities \citep{hinton2015distilling}, while DKD separates target-class and non-target supervision \citep{zhao2022decoupled}. These approaches identify useful teacher signals; our analysis asks which signals remain attainable when existing task discrimination must be preserved.

\noindent\textbf{Theoretical perspectives.} KD theory addresses capacity mismatch \citep{mirzadeh2020improved}, teacher quality and statistical supervision \citep{menon2021statistical}, self-distillation as regularization \citep{mobahi2020self}, and data geometry or optimization-induced deviations \citep{phuong2019understanding,nagarajan2023deviations}. We instead characterize what remains learnable, and at what cost, while preserving the student's acquired class comparisons.

\noindent\textbf{Preservation and gradient coordination.} MPT reduces prediction regressions during model updates through margin calibration and dual-source distillation \citep{ricci2026mitigating}. PCGrad projects conflicting gradients \citep{yu2020gradient}; DeepKD decouples momentum updates and filters non-target knowledge \citep{huang2025deepkd}; DTO-KD balances task and distillation gradients \citep{hayder2026dto}. Our analysis first characterizes blocking and transfer cost in the task-preserving output region. TPKD then keeps the complete label gradient and minimally corrects each example's conditional increment, with tight bounds on retained learning.

\section{What to learn and what to preserve}\label{sec:setup}
\textbf{Learning targets.} Labels specify the correct answer; the teacher also describes relations among alternatives. We distinguish label fitting from two teacher-learning objectives: reproducing the full prediction or learning only its conditional structure among incorrect classes.

Consider an example with label $y\in\{1,\ldots,K\}$, $K\ge3$. The student produces logits $z\in\R^K$ and probabilities $p=\softmax(z)$, where $p_j=e^{z_j}/\sum_{k=1}^Ke^{z_k}$; the teacher supplies probabilities $t$. Both distributions are strictly positive. Define their correct-class confidences $\rho=p_y$, $\tau=t_y$, and conditional probabilities $r_j=p_j/(1-\rho)$, $q_j=t_j/(1-\tau)$ for $j\ne y$. The vectors $r,q$ each sum to one over incorrect classes. With coordinate $y$ first and the other class indices aligned,
$p=(\rho,(1-\rho)r)$, $t=(\tau,(1-\tau)q)$.
For positive distributions $a,b$, let $\KL(a\Vert b)=\sum_j a_j\log(a_j/b_j)$; $\Ber(\tau\Vert\rho)$ denotes the KL between $(\tau,1-\tau)$ and $(\rho,1-\rho)$. Label fitting minimizes $\CE(p,y)=-\log\rho$, full imitation minimizes $L_T(p)=\KL(t\Vert p)$, and conditional learning minimizes $\Shape(p)=\KL(q\Vert r)$. The teacher loss decomposes as \citep{zhao2022decoupled}
\begin{equation}\label{eq:kldecomp}
 L_T(p)=\Ber(\tau\Vert\rho)+(1-\tau)\Shape(p).
\end{equation}
Full imitation matches both teacher confidence and conditional structure. Increasing $\rho$ with $r$ fixed improves label fitting without changing these relations; label progress alone therefore does not measure conditional learning. We use natural logarithms and Euclidean inner products, norms and projections. Logit arguments mean evaluation at $p=\softmax(z)$, with $y,t$ fixed.

\textbf{Preserving learned comparisons.} The student may change its prediction, but should not give back its acquired advantage over any competitor. For each $j\ne y$, define the margin $M_j(p)=\log(p_y/p_j)=z_y-z_j$. Let $\Delta_m^\circ=\{a\in\R^m:a_j>0,\ \sum_{j=1}^ma_j=1\}$ be the positive probability simplex. From reference $p$, the task-preserving predictions are
\begin{equation}\label{eq:safeset}
 \mathcal S(p)=\{p'\in\Delta_K^\circ:M_j(p')\ge M_j(p)\text{ for all }j\ne y\}.
\end{equation}
This protects individual comparisons as well as the label loss. For nonnegative competitor weights $\omega=(\omega_j)_{j\ne y}$, define $L_\omega(z,y)=\log(1+\sum_{j\ne y}\omega_je^{-M_j(z)})$. Preserving all margins is equivalent to not increasing any $L_\omega$: unit weights give CE; a single unit weight isolates one competitor (Appendix~\ref{app:weightedproof}). CE alone can improve while one comparison worsens; Appendix~\ref{app:ce-only} gives its constrained imitation optimum.

\textbf{How learning and protection interact.} Since $p_j=(1-\rho)r_j$, we have $M_j(p)=\log[\rho/(1-\rho)]-\log r_j$. Matching a different conditional target raises some $r_j$, shrinking its margin at fixed $\rho$. Increasing correct-class confidence can compensate. Conditional learning leaves this confidence free; full imitation also seeks the teacher's confidence. We next determine which goal remains attainable within $\mathcal S(p)$, and at what minimum cost.

\section{When imitation stops, what can the teacher still teach?}\label{sec:separation}
\subsection{Why full imitation can become blocked}
Our first result identifies when closer imitation must undo task progress. Once the student has suppressed every incorrect class at least as strongly as the teacher, its current prediction is already the best full imitation permitted by task preservation.
\begin{theorem}[Full-imitation blocking]\label{thm:block}
For positive $p,t$, the inequalities $p_j\le t_j$ for all $j\ne y$ hold if and only if $p$ uniquely minimizes $L_T$ over $\mathcal S(p)$. For any $p'\in\mathcal S(p)$, define $\Delta M_j=M_j(p')-M_j(p)$. Under the blocking condition $p_j\le t_j$ for all $j\ne y$,
\begin{equation}\label{eq:block}
 L_T(p')-L_T(p)=\KL(p\Vert p')+\sum_{j\ne y}(t_j-p_j)\Delta M_j
 \ge\KL(p\Vert p').
\end{equation}
\end{theorem}
The teacher asks for at least as much probability on each incorrect class, while protection permits only nonnegative margin gains. Every term on the right of Eq.~\eqref{eq:block} is therefore nonnegative: any change from $p$ strictly increases full imitation error. Yet conditional knowledge can remain unlearned ($r\ne q$) throughout a nonempty region of blocked states (Appendix~\ref{app:blockproof}).

\subsection{Learning the remaining structure, and its price}
\textbf{Blocking full imitation does not exhaust the teacher's knowledge.} The student can still match the teacher's relative probabilities among incorrect classes. Raising a competitor's conditional probability requires enough extra correct-class confidence to preserve its margin. The next result gives the minimum increase needed for complete transfer.

For a candidate confidence $a\in(0,1)$, let $p'=(a,(1-a)q)$ match the full conditional target. Define $\logit(a)=\log[a/(1-a)]$, the log-odds of the correct class, $\mathcal R_\infty=\max_{j\ne y}q_j/r_j$, and $D_\infty(q\Vert r)=\log\mathcal R_\infty$. Let $a_{\min}$ be the smallest confidence permitting task-preserving transfer.
\begin{theorem}[Minimum compensation for exact transfer]\label{thm:cost}
The prediction $p'$ belongs to $\mathcal S(p)$ if and only if
\begin{equation}\label{eq:mincost}
 \logit(a)-\logit(\rho)\ge D_\infty(q\Vert r),
 \qquad a_{\min}=\frac{\rho\mathcal R_\infty}{1-\rho+\rho\mathcal R_\infty}<1.
\end{equation}
A continuous path attains $a_{\min}$ while every margin remains nondecreasing. If $r\ne q$, then $a_{\min}>\rho$ and conditional KL decreases strictly along the path to zero.
\end{theorem}
The most underestimated competitor sets the price. To attain it, for $s\in[0,1]$ define $r(s)=(1-s)r+sq$, choose $a(s)$ by $\logit a(s)=\logit\rho+\log(1-s+s\mathcal R_\infty)$, and set $p(s)=(a(s),(1-a(s))r(s))$. This moves the whole conditional distribution toward the teacher while compensating just enough at each point (Appendices~\ref{app:costproof}--\ref{app:pathproof}); Appendix~\ref{app:frontier} gives optimal partial transfer at smaller budgets.

\begin{figure}[!htbp]
\centering
\begin{minipage}[t]{.485\linewidth}\centering\includegraphics[width=\linewidth]{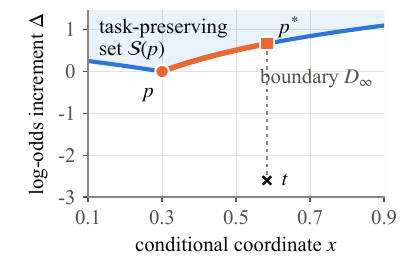}\end{minipage}\hfill
\begin{minipage}[t]{.485\linewidth}\centering\includegraphics[width=\linewidth]{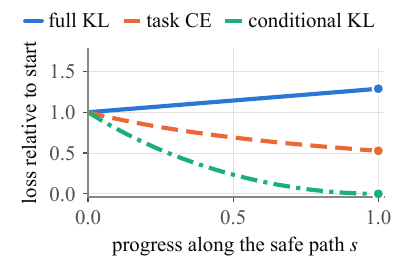}\end{minipage}
\caption{\textbf{Full imitation can worsen while conditional knowledge is learned.} The safe path attains the minimum confidence $a_{\min}\approx0.946$. Label CE and conditional KL fall while full KL rises; losses are normalized by their initial values.}\label{fig:feasibility}
\end{figure}

\textbf{A concrete example.} In Figure~\ref{fig:feasibility}, $y=1$, $p=(0.90,0.07,0.03)$ and $t=(0.40,0.25,0.35)$. Both incorrect-class probabilities are below the teacher's, so full imitation is blocked. Yet the student favors class 2 over class 3 ($7{:}3$), whereas the teacher favors class 3 ($5{:}7$). At the transfer endpoint $p^*=(a_{\min},(1-a_{\min})q)$, $a_{\min}=35/37\approx0.946$: the class-3 margin stays fixed and the class-2 margin grows. The plot uses conditional coordinate $x=r_3$ and confidence increment $\Delta=\logit(a)-\logit(\rho)$. Appendix~\ref{app:pathproof} works through the calculation.

\textbf{The separation is strongest near perfect label fitting.} For $\varepsilon\in(0,1)$, define $p^{(\varepsilon)}=(1-\varepsilon,\varepsilon q)$, whose conditional distribution is $r^{(\varepsilon)}=q$. Then, as $\varepsilon\downarrow0$,
\begin{equation}\label{eq:separation}
 \CE(p^{(\varepsilon)},y)\to0,\qquad \KL(q\Vert r^{(\varepsilon)})=0,
 \qquad \KL(t\Vert p^{(\varepsilon)})\to\infty.
\end{equation}
Learning the label and the teacher's conditional structure therefore need not make the student a closer full imitator. Theorem~\ref{thm:cost} identifies the minimum total confidence compensation required to complete this transfer. \textbf{We now turn to the local learning problem: how can each update acquire conditional knowledge while preserving all the progress of the corresponding label step?} TPKD addresses this question by making the smallest necessary correction to the combined learning direction.

\section{Task-preserving knowledge distillation}\label{sec:tpkd}
\subsection{Keep label learning intact and correct the teacher signal}\label{sec:core}
The separation suggests a simple design: keep the progress of ordinary label learning and correct only the additional teacher signal that would interfere with it. The correction should be minimal, so protection does not unnecessarily discard teacher knowledge.

Let $e_j\in\R^K$ have one in coordinate $j$ and zeros elsewhere. The two logit gradients are $h=\nabla_z\CE=p-e_y$ and $u=\nabla_z\Shape=(0,r-q)$, where the zero occupies coordinate $y$. Let $\one\in\R^K$ denote the all-ones vector. Conditional gradients lie in the subspace $\Ls$. Directions that can be subtracted without reducing any margin form the safe cone $\Kc$:
\[
 \Ls=\{x\in\R^K:x_y=0,\ \one^\top x=0\},\qquad
 \Kc=\{x\in\R^K:x_j-x_y\ge0\text{ for all }j\ne y\}.
\]
A nonzero direction in $\Ls$ cannot belong to $\Kc$: changing only the relative wrong-class scores must favor some competitor. Thus $\Ls\cap\Kc=\{0\}$. Write $\Pi_{\Kc}$ for Euclidean projection onto $\Kc$, let $d$ be the corrected conditional direction and $v$ the complete update direction. With output step size $\eta\ge0$, TPKD uses
\begin{equation}\label{eq:core}
 d=\Pi_{\Kc}(u),\qquad \boxed{v=h+\tfrac12d,\qquad z^+=z-\eta v.}
\end{equation}
Equivalently, $v$ is the nearest direction to $h+u/2$ that preserves every label-step margin gain. The projection costs $O(K\log K)$ per example (Appendix~\ref{app:projectionproof}).

\textbf{The reference is the progress that the label step would have achieved on its own.} Starting from the same logits and using the same step size, adding the corrected conditional signal preserves every margin gain of that label step:
\begin{equation}\label{eq:marginorder}
 \begin{aligned}
 M_j(z-\eta v)&\ge M_j(z-\eta h)\ge M_j(z),&&j\ne y,\\
 \CE(z-\eta v)&\le\CE(z-\eta h)\le\CE(z).
 \end{aligned}
\end{equation}
The same ordering holds for every weighted label loss $L_\omega$ with $\omega\ge0$. Thus TPKD preserves not only the student's existing discrimination, but also the additional discrimination that the corresponding CE step would have gained.

\begin{figure}[!htbp]
\centering
\begin{tikzpicture}[font=\normalsize,>=Latex,
 box/.style={draw=black!65,rounded corners=2pt,align=center,minimum height=.85cm,text width=3.5cm,inner sep=4pt},
 arr/.style={-{Latex[length=2mm]},draw=black!65,line width=.8pt}]
\node[box] (input) at (0,0) {Student logits $z$,\\label $y$};
\node[box] (h) at (4.65,0) {Label signal\\$h=p-e_y$};
\node[box] (target) at (0,-1.12) {Teacher/student\\non-target logits};
\node[box] (u) at (4.65,-1.12) {Conditional signal\\$u=(0,r-q)$};
\node[box,fill=black!4] (d) at (9.3,-1.12) {Safe correction\\$d=\Pi_{\Kc}(u)$};
\node[box,fill=black!7] (v) at (9.3,0) {\textbf{Keep both signals}\\$v=h+d/2$};
\draw[arr](input)--(h);\draw[arr](h)--(v);\draw[arr](target)--(u);\draw[arr](u)--(d);\draw[arr](d)--(v);
\end{tikzpicture}
\caption{\textbf{TPKD keeps the label signal intact.} Only the teacher-conditional direction is corrected. The combined direction is passed to ordinary backpropagation through Eq.~\eqref{eq:surrogate}.}\label{fig:training-workflow}
\end{figure}
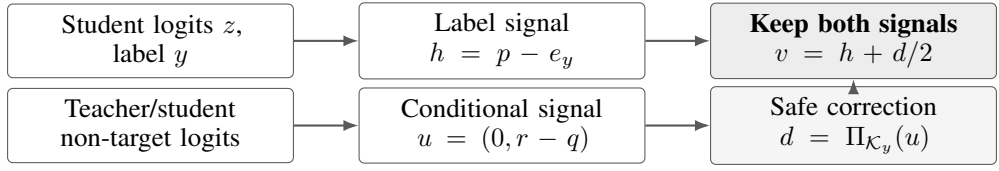

\textbf{Training uses ordinary backpropagation.} Let $\theta\in\R^P$ collect the $P$ network parameters; subscript $i$ indexes the $N$ batch examples. Define the output Jacobian $J_i=\partial z_i/\partial\theta\in\R^{K\times P}$. Holding $v_i$ fixed during differentiation with $\stopgrad$, we inject it through the surrogate loss
\begin{equation}\label{eq:surrogate}
 L_{\rm sur}=\frac1N\sum_{i=1}^N\ip{\stopgrad(v_i)}{z_i},\qquad
 \nabla_\theta L_{\rm sur}=\frac1N\sum_{i=1}^NJ_i^\top v_i.
\end{equation}
For teacher logits $z^T$, let $z^T_{\neg y}$ denote the non-target coordinates. The fixed teacher supplies its original $q=\softmax(z^T_{\neg y})$. Figure~\ref{fig:training-workflow} summarizes the two learning signals and their combination.

\subsection{Protection retains real conditional learning}\label{sec:clipped}\label{sec:retention}
\textbf{What does the correction change?} It caps the teacher's strongest requests to raise competing classes and turns the clipped mass into correct-class compensation. Let $\widetilde q$ be the retained teacher masses and $\ell=-d_y\ge0$ the compensation. With $[a]_+=\max(a,0)$, $\ell$ uniquely solves $\ell=\sum_{j\ne y}[q_j-r_j-\ell]_+$, giving
\begin{equation}\label{eq:clipped-core}
 \widetilde q_j=\min(q_j,r_j+\ell),\qquad
 d=(-\ell,r-\widetilde q),\qquad \sum_{j\ne y}\widetilde q_j=1-\ell.
\end{equation}
Freezing the clipping at the current output gives a local loss whose gradient equals $d$ at that output: fit the normalized clipped teacher and increase correct-class log-odds (Appendix~\ref{app:local-surrogate}). The extra margin over the CE endpoint is $\frac\eta2[r_j-q_j+\ell]_+$: the most demanding competitors keep exactly the CE margin; the others gain more.

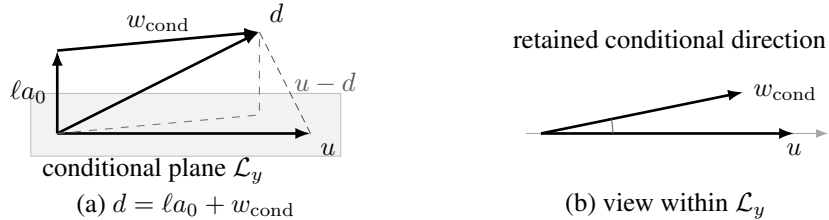
\begin{figure}[!htbp]
\centering
\begin{tikzpicture}[font=\normalsize,>=Latex,
vec/.style={-{Latex[length=2mm]},line width=1pt}]
\begin{scope}
\fill[black!5] (-.35,-.30) rectangle (3.75,.53);
\draw[black!25] (-.35,-.30) rectangle (3.75,.53);
\node[anchor=west] at (-.32,-.49) {conditional plane $\mathcal L_y$};
\coordinate (O) at (0,0);\coordinate (U) at (3.35,0);
\coordinate (W) at (2.674597,0.245665);\coordinate (C) at (0.000000,1.096791);\coordinate (D) at (2.674597,1.342455);
\draw[vec] (O)--(U) node[below right] {$u$};
\draw[vec] (O)--(D) node[above right] {$d$};
\draw[vec] (O)--(C) node[midway,left] {$\ell a_0$};
\draw[vec] (C)--(D) node[midway,above] {$w_{\rm cond}$};
\draw[dashed,black!55] (O)--(W)--(D);
\draw[densely dashed,black!65] (D)--(U) node[midway,right] {$u-d$};
\node at (1.7,-.95) {(a) $d=\ell a_0+w_{\rm cond}$};
\end{scope}
\begin{scope}[xshift=6.4cm]
\draw[black!35,-{Latex[length=1.6mm]}] (-.2,0)--(3.8,0);
\draw[vec] (0,0)--(3.35,0) node[below] {$u$};
\draw[vec] (0,0)--(2.674597,0.545922) node[right] {$w_{\rm cond}$};
\draw[black!60] (.95,0) arc[start angle=0,end angle=11.536386237016446,radius=.95];
\node at (1.70,1.25) {retained conditional direction};

\node at (1.65,-.95) {(b) view within $\mathcal L_y$};
\end{scope}
\end{tikzpicture}
\caption{\textbf{Protection preserves a genuine conditional-learning direction.} The corrected direction combines confidence compensation with a pure conditional component (a), which remains closely aligned with the original signal (b). The four-class example is constructed in Appendix~\ref{app:geometry-example}.}\label{fig:projection-geometry}
\end{figure}

\textbf{The important question is whether protection preserves genuine conditional learning or merely increases confidence in the correct class.} To distinguish these effects, define the symmetric confidence direction $a_0=(-1,1/(K-1),\ldots,1/(K-1))$ and the remaining component $w_{\rm cond}=d-\ell a_0$. Then
$d=\ell a_0+w_{\rm cond}$, $w_{\rm cond}\in\Ls$, $w_{\rm cond}\perp a_0$.
A step along $-a_0$ increases every correct-class margin equally and leaves the relative probabilities among incorrect classes unchanged. The component $w_{\rm cond}$ changes those relations. The next theorem shows that this component remains nonzero and aligned with the teacher's conditional signal whenever conditional error remains.
\begin{theorem}[Retained conditional learning]\label{thm:retention}\label{prop:alignment}
For $u\in\Ls$, define $\gamma_K=K/[2(K-1)]$ and $c_K=2\sqrt{\gamma_K}/(1+\gamma_K)$. Then
\begin{align}
 &u^\top d=\|d\|^2\ge\gamma_K\|u\|^2,
 \qquad \|w_{\rm cond}\|^2\ge\gamma_K\|d\|^2\ge\gamma_K^2\|u\|^2,\label{eq:retention}\\
 &\cos\angle(u,w_{\rm cond})\ge c_K\ge\frac{2\sqrt2}{3}\qquad(u\ne0).\label{eq:relation-angle}
\end{align}
Both norm bounds are tight and can be attained simultaneously.
\end{theorem}
Here $u^\top d$ measures first-order conditional loss reduction along $-d$. The first bound retains more than half the reduction along $-u$ at the same step size; the remaining bounds ensure a nonzero, aligned conditional component whenever $r\ne q$ (Figure~\ref{fig:projection-geometry}). Appendix~\ref{app:geometry} quantifies the removed signal and compensation, and establishes optimality under local quadratic and fixed-displacement budgets.

\subsection{Both kinds of learning can be completed together}\label{sec:dynamics}
\textbf{Can the learning retained in each step accumulate until both objectives are achieved?} We prove that it can: for a fixed positive teacher-conditional target and sufficiently small constant output steps, label fitting and conditional matching converge together. To track their joint progress, define the potential $\Phi(z)=4\CE(z,y)+\Shape(z)$ and its logit gradient $g_\Phi=\nabla_z\Phi=4h+u$.
A subscript $k$ denotes evaluation at output $z_k$, for example $p_k=\softmax(z_k)$ and $v_k=v(z_k)$.
\begin{theorem}[Joint descent and completed learning]\label{thm:convergence}\label{prop:full-alignment}
The complete direction obeys the tight alignment bound
\begin{equation}\label{eq:full-alignment}
 g_\Phi^\top v\ge\|v\|^2+\frac{\gamma_K}{4}\|g_\Phi\|^2,
 \qquad \cos\angle(g_\Phi,v)\ge\sqrt{\gamma_K}.
\end{equation}
For fixed positive $q$, finite initial logits, and $z_{k+1}=z_k-\eta v_k$ with $0<\eta\le4/5$,
\begin{equation}\label{eq:descentcertificate}
 \Phi(z_{k+1})\le\Phi(z_k)-\frac{\eta\gamma_K}{4}\|g_{\Phi,k}\|^2
 -\eta(1-5\eta/4)\|v_k\|^2.
\end{equation}
Consequently, $p_{k,y}\to1$, $r_k\to q$, and $v_k\to0$.
\end{theorem}
The retained conditional signal therefore supports completed learning, not just a favorable local direction. Joint and component gradient residuals have vanishing time-averaged squared norms (Appendix~\ref{app:dynamicsproof}); Appendix~\ref{app:conditional-descent} gives conditions for a complete step to lower its own conditional KL. For comparison, define the uniform non-target vector by $\nu_j=1/(K-1)$ for $j\ne y$. CE output gradient flow, or fixed steps $0<\eta\le1/20$, fits the label but drives $r\to\nu$. For nonuniform $q$, both rules fit the label, but TPKD retains the teacher's conditional structure rather than a uniform distribution (Appendix~\ref{app:ceuniform}).

\subsection{Turning output progress into parameter learning}\label{sec:realization}
The output construction becomes useful for training through two connections: exact execution by a classification head, and conditional learning beyond compensation under ordinary backpropagation.

\begin{proposition}[Exact head update]\label{prop:exact}
If the frozen batch feature matrix has full row rank, a unique minimum-Frobenius-norm head displacement realizes Eq.~\eqref{eq:core} exactly for every example. It therefore inherits the step's margin and loss orderings in Eq.~\eqref{eq:marginorder} (Appendix~\ref{app:exactproof}).
\end{proposition}
\textbf{Learning beyond compensation.} Compare the parameter steps from Eq.~\eqref{eq:surrogate} for TPKD and $h_i+\ell_i a_{0,i}/2$: the same confidence compensation without direct conditional adjustment. Backpropagation can distort these signals. Let $\chi_c,\chi_\Phi$ be its largest-to-smallest squared stretch ratios on the spans of the batch-concatenated pairs $(u_i,w_{{\rm cond},i})$ and $(g_{\Phi,i},v_i)$, respectively. Zero minimum stretch gives an infinite ratio. Define $\kappa_\Phi=(1+\sqrt{\gamma_K})/(1-\sqrt{\gamma_K})$ and $\kappa_c=\kappa_\Phi^2$.
\begin{theorem}[Conditional learning through network parameters]\label{thm:parameter}\label{cor:relation-backprop}\label{cor:kernel}
At a differentiable parameter state, for sufficiently small equal positive parameter steps:
\textup{(i)} if some $u_i\ne0$ and $\chi_c<\kappa_c$, TPKD has a nonzero conditional parameter increment over compensation only and achieves strictly lower batch-average conditional KL;
\textup{(ii)} if both joint batch vectors are nonzero and $\chi_\Phi<\kappa_\Phi$, TPKD strictly decreases the batch mean of $\Phi$.
\end{theorem}
The first conclusion guarantees a useful conditional increment, not merely extra confidence; the second guarantees progress of the complete update. These results establish both a parameter implementation and, under the stated conditions, genuine learning from the retained teacher signal. Appendix~\ref{app:parameters} gives the equivalent formulas, proofs, quantitative bounds, computation and measured-angle refinements.

\section{Experiments}\label{sec:experiments}
\label{sec:mechanism}\label{sec:regime-experiment}\label{sec:exact-experiment}\label{sec:backprop-experiment}
The mechanism study uses CIFAR-100 \citep{krizhevsky2009learning}, with a VOLO-D2 teacher and PiT-B student \citep{yuan2023volo,heo2021rethinking}, under the main training protocol. CE20/40/60 denote label-only students after 20/40/60 epochs. Margins measure the correct class's advantage over each competitor; conditional KL measures the error in learning the teacher's relations among incorrect classes.

\noindent\textbf{Blocking in ordinary training.} We first test whether ordinary label training can block full imitation while leaving conditional knowledge unlearned. Each CE state is evaluated on 2,000 fixed images under eight views; an image--view pair is blocked when the student's probability for every incorrect class is no greater than the teacher's. Blocking rises from 50.00\% to 58.90\%, yet every blocked pair has positive conditional error (Table~\ref{tab:states}). Thus the theoretical obstruction occurs in ordinary training while teacher knowledge remains available.

\begin{table}[!htbp]
\centering
\caption{\textbf{Full-imitation blocking leaves conditional knowledge unlearned.} Conditional error and transfer cost are means within the blocked set (nats).}\label{tab:states}
\begin{tabular*}{\linewidth}{@{\extracolsep{\fill}}lrrr@{}}
\toprule
Student state & Blocked (\%) & Remaining conditional KL & Minimum log-odds cost \\
\midrule
CE20 & 50.00 & 1.696 & 7.520 \\
CE40 & 55.29 & 1.766 & 7.632 \\
CE60 & 58.90 & 1.803 & 7.569 \\
\bottomrule
\end{tabular*}
\end{table}

\noindent\textbf{Protection without discarding knowledge.} Exact head updates on frozen features isolate whether the prescribed correction protects the task without discarding conditional learning. At CE20 and CE60, we compare TPKD with CE, compensation only, unprojected learning, and CE plus full KL from the same start. Compensation only keeps the correct-class push but removes changes among incorrect classes. TPKD gains conditional knowledge beyond CE on all 64 batches and all 4,096 example states while preserving every CE-step margin to numerical precision. Compensation only has CE's conditional endpoint; unprojected learning loses at least one CE-step margin on every example. TPKD therefore combines the two desired effects. Minimum retention and complete-update cosine exceed the 100-class bounds of approximately 50.51\% and 0.7107 (Theorems~\ref{thm:retention}--\ref{thm:convergence}); all joint-descent checks pass (Table~\ref{tab:exact}).

\begin{table}[!htbp]
\centering
\caption{\textbf{Exact head steps learn conditional structure while preserving label-step progress.}}\label{tab:exact}
\setlength{\tabcolsep}{3pt}
\begin{tabular*}{\linewidth}{@{\extracolsep{\fill}}lrrrrr@{}}
\toprule
State & Gain ($10^{-3}$) & KL change ($10^{-3}$) & Min. retained (\%) & Min. cosine & $\Phi$ change ($10^{-3}$) \\
\midrule
CE20 & 1.195 & -1.359 & 52.13 & 0.7220 & -3.302 \\
CE60 & 1.213 & -1.283 & 52.13 & 0.7221 & -1.831 \\
\bottomrule
\end{tabular*}
\par\smallskip\raggedright
KL denotes conditional KL. Gain is CE minus TPKD at the endpoint; changes are TPKD endpoint minus common start. Loss columns are means; minima are over individual examples. Retention is $u^\top d/\|u\|^2$; cosine compares $v$ with $g_\Phi$. Each state: $32\times64$ examples; output step $0.01$.
\end{table}

\noindent\textbf{Conditional learning through the network.} We next test whether this conditional increment remains useful under ordinary full-network optimization. At CE20 and CE60, 16 native-optimizer pairs per state compare TPKD with compensation only from identical model, optimizer and random states. For $K=100$, Theorem~\ref{thm:parameter} gives distortion-ratio limits of approximately 34.96 for conditional learning and 5.91 for joint descent. All 32 batches pass both geometric tests, and every native pair has lower conditional KL under TPKD (Table~\ref{tab:certificate}). These paired gains isolate conditional learning beyond compensation.

\begin{table}[!htbp]
\centering
\caption{\textbf{The conditional advantage survives ordinary backpropagation.}}\label{tab:certificate}
\begin{tabular*}{\linewidth}{@{\extracolsep{\fill}}lrrrr@{}}
\toprule
State & Conditional test & Joint descent test & Positive pairs & Mean gain ($10^{-3}$) \\
\midrule
CE20 & 16/16 & 16/16 & 16/16 & 8.252 \\
CE60 & 16/16 & 16/16 & 16/16 & 9.886 \\
\bottomrule
\end{tabular*}
\par\smallskip\raggedright
Gain is compensation-only endpoint conditional KL minus TPKD endpoint conditional KL.
\end{table}

\noindent\textbf{Accumulation over successive updates.} Finally, we test whether the single-step advantage accumulates. From CE60, CE, compensation only and TPKD follow the same data sequence for 128 native-optimizer iterations. On 2,048 fixed observation images separate from the update images, TPKD lowers conditional KL from 1.800 to 1.557. Its endpoint KL is 0.244 nats below CE's and 0.320 below compensation only's: the extra conditional learning persists over successive updates. Protocols are in Appendix~\ref{app:experimental-details}.

\subsection{Full training in vision and text}\label{sec:application-experiment}
\textbf{TPKD is not limited to vision:} its update uses class probabilities and labels, not modality-specific features. We therefore also evaluate full training on CLINC150 text intent classification \citep{larson2019evaluation}, with BERT-large as teacher and BERT-Mini as student \citep{devlin2019bert,turc2019well}. Table~\ref{tab:application} compares CE and nine distillation methods \citep{hinton2015distilling,zhao2022decoupled,roth2024fantastic,yang2025dhkd,hayder2026dto} using shared within-domain protocols and final-epoch evaluation fixed in advance over three seeds (Appendix~\ref{app:setup}).

TPKD reaches 88.05\% on CIFAR-100 and 93.81\% on CLINC150, improving over standard KD by 0.47 and 0.35 percentage points and over CE by 0.68 and 0.56 points. Retained conditional learning thus benefits full training in both domains.
\begin{table}[!htbp]
\centering
\caption{\textbf{Full-training accuracy (\%).} Mean $\pm$ sample standard deviation over three seeds.}\label{tab:application}
\begin{minipage}[t]{.485\linewidth}
\setlength{\tabcolsep}{2pt}
\begin{tabular*}{\linewidth}{@{\extracolsep{\fill}}lcc@{}}
\toprule
Method & CIFAR-100 & CLINC150 \\
 & 60 epochs & 4 epochs \\
\midrule
CE & $87.37 \pm 0.07$ & $93.24 \pm 0.28$ \\
KD & $87.58 \pm 0.07$ & $93.46 \pm 0.34$ \\
KL-Dist & $82.26 \pm 0.05$ & $93.38 \pm 0.19$ \\
DKD & $84.53 \pm 0.09$ & $93.59 \pm 0.20$ \\
DP-U & $83.14 \pm 0.04$ & $93.38 \pm 0.19$ \\
\bottomrule
\end{tabular*}
\end{minipage}\hfill
\begin{minipage}[t]{.485\linewidth}
\setlength{\tabcolsep}{2pt}
\begin{tabular*}{\linewidth}{@{\extracolsep{\fill}}lcc@{}}
\toprule
Method & CIFAR-100 & CLINC150 \\
 & 60 epochs & 4 epochs \\
\midrule
DP-S & $83.65 \pm 0.01$ & $93.44 \pm 0.28$ \\
XE-KL & $84.17 \pm 0.05$ & $93.41 \pm 0.22$ \\
DHKD & $85.83 \pm 0.10$ & $93.29 \pm 0.38$ \\
DTO-KD$^*$ & $87.81 \pm 0.11$ & $93.46 \pm 0.23$ \\
\textbf{TPKD} & {\bfseries\boldmath $88.05 \pm 0.12$} & {\bfseries\boldmath $93.81 \pm 0.27$} \\
\bottomrule
\end{tabular*}
\end{minipage}
\par\smallskip\raggedright
$^*$Text DTO-KD includes multi-layer feature distillation. Full configurations are in Appendix~\ref{app:setup}.
\end{table}

\subsection{Which components make the difference?}\label{sec:ablation}
\textbf{Independent label learning.} Matching the conditional target makes $d=0$, but the CE gradient continues fitting the label. Removing it lowers accuracy by 2.72 points in vision and 2.21 in text (Table~\ref{tab:ablation}).

\textbf{Learning the right teacher object.} Safe full KL keeps CE and projection but uses $p-t$ instead of the conditional gradient. In a blocked state, its projected teacher increment is zero, whereas TPKD's remains nonzero when $r\ne q$ (Appendix~\ref{app:ablationproof}). Accuracy falls to 87.21\% and 93.33\%, close to CE. Protecting full imitation alone does not recover the benefit of conditional learning.

\textbf{The cost of protection.} Without projection, accuracy is 87.83\% and 93.83\%: TPKD is 0.22 points higher in vision and differs by only 0.02 points in text. The correction thus provides the demonstrated margin protection while retaining closely matched predictive performance.
\begin{table}[!htbp]
\centering
\caption{\textbf{Component ablations.} Accuracy (\%), mean $\pm$ sample standard deviation over three seeds.}\label{tab:ablation}
\begin{tabular*}{\linewidth}{@{\extracolsep{\fill}}llcc@{}}
\toprule
Condition & Direction & CIFAR-100 & CLINC150 \\
\midrule
CE & $h$ & $87.37 \pm 0.07$ & $93.24 \pm 0.28$ \\
Without CE gradient & $d/2$ & $85.33 \pm 0.05$ & $91.60 \pm 0.07$ \\
Safe full-KL & $h+\Pi_{\Kc}(p-t)/2$ & $87.21 \pm 0.04$ & $93.33 \pm 0.29$ \\
Without projection & $h+u/2$ & $87.83 \pm 0.03$ & $93.83 \pm 0.23$ \\
\textbf{TPKD} & $h+d/2$ & $88.05 \pm 0.12$ & $93.81 \pm 0.27$ \\
\bottomrule
\end{tabular*}
\end{table}

\section{Conclusion}\label{sec:conclusion}
Full imitation can become blocked before the teacher's knowledge is exhausted. We prove that its conditional structure remains fully transferable under task preservation and determine the exact confidence compensation required. TPKD implements this separation by retaining the label gradient and adding the nearest safe conditional direction. Tight retention and joint-learning results explain why genuine conditional learning survives; parameter-space results connect it to network updates. Mechanism experiments locate blocking in ordinary training and follow conditional gains through successive updates, while full training and ablations demonstrate their contribution in vision and text. Learning from a teacher need not mean reproducing its full prediction: conditional knowledge can be acquired without surrendering label-learning progress.

\label{sec:mainend}
\clearpage
\subsection*{AI use statement}
No generative AI tools were used in conducting this research or preparing this manuscript.

\subsubsection*{Reproducibility statement}
Appendices~\ref{app:proofs}--\ref{app:parameters} prove the results stated in the main text; Appendix~\ref{app:experimental-details} specifies the reported experiments. The source bundle includes the reported accuracy summaries, table-generation scripts and analytical checks.
\bibliography{references}
\bibliographystyle{iclr2027_conference}
\clearpage
\appendix
\numberwithin{equation}{section}
\section{Proofs supporting task preservation and the separation}\label{app:proofs}
This appendix proves the equivalence in Section~\ref{sec:setup}, Theorems~\ref{thm:block}--\ref{thm:cost}, and the two component consequences stated in Section~\ref{sec:ablation}.
\subsection{Weighted task losses and per-margin preservation}\label{app:weightedproof}
For nonnegative competitor weights $\omega$, let $L_\omega(z,y)=\log(1+\sum_{j\ne y}\omega_j e^{-M_j(z)})$. The all-ones choice is cross-entropy; a single nonzero weight isolates one competitor.
\paragraph{Proof of the equivalence in Section~\ref{sec:setup}.}
For logits $z,z'\in\R^K$, if $M_j(z')\ge M_j(z)$ for all $j\ne y$, then $e^{-M_j(z')}\le e^{-M_j(z)}$; any $\omega_j\ge0$ preserves the termwise inequality, and summing and using the monotonicity of $\log$ gives $L_\omega(z',y)\le L_\omega(z,y)$. Conversely, for any $j\ne y$ take $\omega_j=1$ and all other weights zero, which gives $\log(1+e^{-M_j(z')})\le\log(1+e^{-M_j(z)})$; since $M\mapsto\log(1+e^{-M})$ is strictly decreasing, $M_j(z')\ge M_j(z)$. Taking $z'$ and the reference to be $z-\eta v$ and $z-\eta h$ yields the ordering of \tpkd\ against the label base under every competitor weighting.

\subsection{Proof of Theorem~\ref{thm:block}}\label{app:blockproof}
Let $\Delta M_j=M_j(p')-M_j(p)$. Since $\sum_j(t_j-p_j)=0$, the common quantity $\log(p_y/p'_y)$ may be subtracted from every term without changing the sum, so
\begin{align}
 \KL(t\Vert p')-\KL(t\Vert p)-\KL(p\Vert p')
 &=\sum_j(t_j-p_j)\log\frac{p_j}{p'_j}\notag\\
 &=\sum_{j\ne y}(t_j-p_j)\Bigl[\log\frac{p_j}{p'_j}-\log\frac{p_y}{p'_y}\Bigr]
 =\sum_{j\ne y}(t_j-p_j)\,\Delta M_j .
\end{align}
This gives the identity in Eq.~\eqref{eq:block} before imposing the sign conditions. When $p'\in\mathcal S(p)$ and $p_j\le t_j$ for all non-target coordinates, the right-hand side is nonnegative; for strictly positive probabilities $\KL(p\Vert p')$ is strictly positive when $p'\ne p$, so $p$ is the unique minimizer and Eq.~\eqref{eq:block} follows. Conversely, if $p_j>t_j$ for some $j$, let $z'=z-\eta e_j$ and $p'=\softmax(z')$: this leaves the other margins unchanged and increases the $j$-th, so $p'\in\mathcal S(p)$; since $\partial_{z_j}\KL(t\Vert\softmax(z))=p_j-t_j$, the derivative of the full KL in $\eta$ at zero equals $-(p_j-t_j)<0$, and $p$ is not a minimizer.

\paragraph{Strict blocking with residual conditional error.}
For any fixed positive $r\ne q$ and $\tau<1$, every $\rho$ with
\[0<1-\rho<(1-\tau)\min_{j\ne y}q_j/r_j\]
is strictly blocked and has $\KL(q\Vert r)>0$. Thus blocking with unlearned conditional structure has nonempty interior.

\subsection{Proof of Theorem~\ref{thm:cost}}\label{app:costproof}
With $\mathcal R_\infty=\max_{j\ne y}q_j/r_j$, the smallest feasible correct-class probability is
\begin{equation}
 a_{\min}=\frac{\rho\,\mathcal R_\infty}{1-\rho+\rho\,\mathcal R_\infty}<1.
\end{equation}
For $p'=(a,(1-a)q)$, the $j$-th margin constraint $M_j(p')\ge M_j(p)$ is equivalent to
\[
 \log\frac{a}{(1-a)q_j}\ge\log\frac{\rho}{(1-\rho)r_j}
 \iff\logit(a)-\logit(\rho)\ge\log\frac{q_j}{r_j}.
\]
Taking the maximum over $j$ gives $D_\infty(q\Vert r)$; solving $\logit(a)=\logit(\rho)+\log\mathcal R_\infty$ for $a$ gives Eq.~\eqref{eq:mincost}. Normalization $\sum_jq_j=\sum_jr_j=1$ implies $\mathcal R_\infty\ge1$ with equality if and only if $q=r$; positivity ensures $\mathcal R_\infty<\infty$, hence $a_{\min}<1$.

\subsection{The minimum-compensation path}\label{app:pathproof}
For $s\in[0,1]$, define
\begin{equation}\label{eq:path}
 r(s)=(1-s)r+sq,\qquad
 a(s)=\frac{\rho(1-s+s\mathcal R_\infty)}{1-\rho+\rho(1-s+s\mathcal R_\infty)} .
\end{equation}
\paragraph{Proof.}
Write $l_j=q_j/r_j\le\mathcal R_\infty$. By Eq.~\eqref{eq:path}, $r_j(s)/r_j=1-s+sl_j$ and $\logit a(s)-\logit\rho=\log(1-s+s\mathcal R_\infty)$, so
\begin{align}
 M_j(p(s))-M_j(p)&=\log(1-s+s\mathcal R_\infty)-\log(1-s+sl_j)\ \ge0,\\
 \frac{d}{ds}M_j(p(s))&=\frac{\mathcal R_\infty-l_j}{(1-s+s\mathcal R_\infty)(1-s+sl_j)}\ \ge0 .
\end{align}
With $\Delta_j=q_j-r_j$ we have $q_j=r_j(s)+(1-s)\Delta_j$ and $\sum_j\Delta_j=0$, hence
\begin{align}
 \frac{d}{ds}\KL(q\Vert r(s))
 =-\sum_j\frac{q_j\Delta_j}{r_j(s)}
 =-(1-s)\sum_j\frac{\Delta_j^2}{r_j(s)} .
\end{align}
If $q\ne r$ this derivative is strictly negative for $s<1$; at $s=1$, $r(1)=q$ and $a(1)=a_{\min}$.

\paragraph{The numerical illustration in Figure~\ref{fig:feasibility}.}
The correct label is class 1, with $p=(0.90,0.07,0.03)$ and $t=(0.40,0.25,0.35)$. Since $0.07<0.25$ and $0.03<0.35$, Theorem~\ref{thm:block} applies. The conditional distributions are nevertheless different:
\[
 \rho=\frac9{10},\quad \tau=\frac25,\qquad
 r=\left(\frac7{10},\frac3{10}\right),\qquad
 q=\left(\frac5{12},\frac7{12}\right).
\]
The largest relative mismatch is on class 3:
\[
 \mathcal R_\infty=\max\left\{\frac{25}{42},\frac{35}{18}\right\}=\frac{35}{18},
 \qquad D_\infty(q\Vert r)=\log\frac{35}{18}\approx0.664976.
\]
Inserting this value into Eq.~\eqref{eq:mincost} gives
\[
 a_{\min}=\frac{(9/10)(35/18)}{1/10+(9/10)(35/18)}=\frac{35}{37},\qquad
 p^*=\left(\frac{35}{37},\frac5{222},\frac7{222}\right).
\]
The endpoint's non-target ratio is $5{:}7$, exactly the teacher's. Its correct-versus-incorrect probability ratios are
\[
 \frac{p_1}{p_2}=\frac{90}{7}\ \longrightarrow\ \frac{p_1^*}{p_2^*}=42,
 \qquad
 \frac{p_1}{p_3}=30\ \longrightarrow\ \frac{p_1^*}{p_3^*}=30.
\]
Thus the class-2 margin increases and the class-3 margin is unchanged. Along the path,
\[
 r_2(s)=\frac7{10}-\frac{17s}{60},\quad
 r_3(s)=\frac3{10}+\frac{17s}{60},\qquad
 a(s)=\frac{\frac9{10}(1+\frac{17s}{18})}{\frac1{10}+\frac9{10}(1+\frac{17s}{18})}.
\]
The left plot uses $x=r_3(s)$ and $\Delta=\logit(a(s))-\logit(\rho)$. For an arbitrary positive conditional vector $(1-x,x)$, its safe boundary is $\Delta=D_\infty((1-x,x)\Vert r)$. The teacher is at $x=7/12$ and $\Delta=\logit(2/5)-\logit(9/10)=\log(2/27)\approx-2.60269$, below this region.

The right plot divides each loss by its own initial value. Substitution into the loss definitions gives
\[
 \begin{array}{c|cc}
 &p&p^*\\\hline
 \CE&-\log(9/10)&-\log(35/37)\\
 \Shape&\frac5{12}\log\frac{25}{42}+\frac7{12}\log\frac{35}{18}&0\\
 L_T&\frac25\log\frac49+\frac14\log\frac{25}{7}+\frac7{20}\log\frac{35}{3}
 &\frac25\log\frac{74}{175}+\frac35\log\frac{111}{10}
 \end{array}
\]
Numerically, label CE changes from approximately $0.105361$ to $0.055570$, conditional KL from $0.171739$ to zero, and full teacher KL from $0.853727$ to $1.099879$. These are evaluations of the analytic path, whose margin monotonicity and conditional descent were established above.

\subsection{The two component consequences used in the ablation}\label{app:ablationproof}
For the control without $h$, $r=q$ implies $u=(0,r-q)=0$ and hence $d=\Pi_{\Kc}(0)=0$. This need not imply that $p_y=1$, because $r$ is independent of the correct-class probability. The ordinary label gradient has $h_y=p_y-1$, so it remains nonzero at every positive prediction with $p_y<1$.

For the safe full-KL control, define the polar cone $\Kc^\circ=\{g\in\R^K:g^\top x\le0\text{ for every }x\in\Kc\}$. Put $g=p-t$ and suppose $p_j\le t_j$ for $j\ne y$. For every $x\in\Kc$, using $\one^\top g=0$,
\[
 g^\top x=\sum_{j\ne y}(t_j-p_j)(x_y-x_j)\le0.
\]
Thus $g$ belongs to the polar cone $\Kc^\circ$ and $\Pi_{\Kc}(g)=0$. If $r\ne q$, Theorem~\ref{thm:retention} instead gives $\|\Pi_{\Kc}(u)\|^2\ge\gamma_K\|u\|^2>0$. This proves the loss of the full-KL teacher increment at a blocked state, while the conditional increment remains available.

\subsection{The imitation optimum when only CE is protected}\label{app:ce-only}
The comparison in Section~\ref{sec:setup} can be solved exactly. Preserving only the label cross-entropy requires $p'_y\ge\rho$. Write $p'=(a,(1-a)x)$, where $x$ is a conditional probability vector. The KL decomposition gives
\[
 \KL(t\Vert p')=\Ber(\tau\Vert a)+(1-\tau)\KL(q\Vert x).
\]
For any feasible $a$, the second term is uniquely minimized by $x=q$. Since
\[
 \frac{\partial}{\partial a}\Ber(\tau\Vert a)=\frac{a-\tau}{a(1-a)},
\]
the unique optimum over $a\ge\rho$ is $a_* =\max(\rho,\tau)$. Thus
\begin{equation}\label{eq:ce-only-optimum}
 p^*_{\rm CE}=(a_*,(1-a_*)q),\qquad
 \min_{p'_y\ge\rho}\KL(t\Vert p')=\Ber(\tau\Vert a_*).
\end{equation}
This solution matches the conditional target without lowering the correct-class probability. It can still reduce an individual competitor margin, which the stronger requirement in Eq.~\eqref{eq:safeset} preserves. For the example in Figure~\ref{fig:feasibility}, $p^*_{\rm CE}=(0.9,1/24,7/120)$ has unchanged CE, but its third-class probability exceeds the initial $0.03$ and hence its third-class margin is smaller.

\subsection{The finite-compensation frontier}\label{app:frontier}
Theorem~\ref{thm:cost} gives the budget needed for complete conditional transfer. With a smaller budget, the best partial transfer also has a closed form. Fix $B\ge0$ as the maximum allowed increase in correct-class log-odds. The attainable conditional distributions are
\begin{equation}\label{eq:budget-feasible}
 \mathcal Q_B(r)=\left\{x\in\R_{>0}^{K-1}:\sum_{j\ne y}x_j=1,\quad
 x_j\le e^B r_j\ (j\ne y)\right\}.
\end{equation}
Indeed, a prediction $(a,(1-a)x)$ preserves the $j$th margin exactly when $x_j/r_j\le\exp(\logit(a)-\logit(\rho))$. Any $x$ in Eq.~\eqref{eq:budget-feasible} is attained safely with $\logit(a)=\logit(\rho)+B$.

Define the least remaining conditional error by
\[
 \mathcal V(B)=\min_{x\in\mathcal Q_B(r)}\KL(q\Vert x).
\]
Let $\lambda>0$ enforce normalization. The unique minimizer $x^*$ is positive and satisfies
\begin{equation}\label{eq:budget-optimum}
 x_j^*=\min\{e^B r_j,q_j/\lambda\},\qquad
 \sum_{j\ne y}x_j^*=1,
\end{equation}
Moreover,
\begin{equation}\label{eq:budget-endpoints}
 \mathcal V(0)=\KL(q\Vert r),\qquad
 \mathcal V(B)=0\quad\Longleftrightarrow\quad B\ge D_\infty(q\Vert r).
\end{equation}
The function $\mathcal V$ is nonincreasing and convex in $B$.

\paragraph{Proof.}
The closure of the feasible set is compact and contains the positive vector $r$. Since $q$ is positive, the objective diverges when any coordinate approaches zero; a positive minimizer therefore exists. Strict convexity of $-\sum_jq_j\log x_j$ gives uniqueness. For $B>0$, let $\mu_j\ge0$ be the multipliers for $x_j\le e^B r_j$. The KKT conditions are
\[
 -q_j/x_j+\lambda+\mu_j=0,\qquad
 \mu_j\ge0,\qquad \mu_j(x_j-e^B r_j)=0.
\]
At least one coordinate is uncapped because the caps sum to $e^B>1$; hence $\lambda=q_j/x_j>0$ there. Uncapped coordinates equal $q_j/\lambda$, while capped coordinates equal $e^B r_j$, yielding Eq.~\eqref{eq:budget-optimum}. At $B=0$ the only feasible vector is $r$; any $0<\lambda\le\min_jq_j/r_j$ gives the same formula. Equation~\eqref{eq:budget-endpoints} follows from $\KL(q\Vert x)=0$ exactly at $x=q$ and the feasibility of that vector.

The sets $\mathcal Q_B(r)$ grow with $B$, so $\mathcal V$ is nonincreasing. For convexity, define $Y=(Y_j)_{j\ne y}$ by $Y_j=\log(x_j/r_j)$ and use the equivalent problem
\begin{equation}\label{eq:budget-log-form}
 \mathcal V(B)=\min_Y\left\{\KL(q\Vert r)-q^\top Y:
 \sum_jr_je^{Y_j}\le1,\quad Y_j\le B\right\}.
\end{equation}
The normalization constraint is active at an optimum: otherwise a coordinate below its cap can be increased, strictly reducing the objective, since $q_j>0$ and the caps sum to at least one. The objective is affine and the constraints are jointly convex in $(Y,B)$. A convex combination of feasible minimizers at two budgets is feasible at their averaged budget and has the averaged objective, proving convexity of $\mathcal V$.

\section{Proofs supporting the safe correction and retained learning}\label{app:geometry}
These proofs support the update in Section~\ref{sec:core}, the clipped representation in Section~\ref{sec:clipped}, and Theorem~\ref{thm:retention}. We use Euclidean projection and its Moreau decomposition \citep{parikh2014proximal}.

\subsection{The projection and task ordering}\label{app:projectionproof}
The general star-cone projection is useful for both TPKD and safe full KL. For an input $x\in\R^K$, sort its non-target coordinates as $x_{(1)}\le\cdots\le x_{(K-1)}$. With an empty prefix sum for $k=0$,
\begin{equation}\label{eq:closedproj}
 \zeta=\min_{0\le k\le K-1}\frac{x_y+\sum_{j=1}^k x_{(j)}}{k+1},
 \qquad (\Pi_{\Kc}x)_y=\zeta,\quad (\Pi_{\Kc}x)_j=\max(x_j,\zeta).
\end{equation}
For TPKD, $x=u$ has $u_y=0$, so $\zeta\le0$. Computing the sorted prefixes costs $O(K\log K)$.
To prove the formula, fix the projected target coordinate at $\zeta$. The other optimal coordinates are $\max(x_j,\zeta)$, reducing the problem to
\[
 F(\zeta)=\tfrac12(\zeta-x_y)^2+\tfrac12\sum_{j\ne y}[\zeta-x_j]_+^2.
\]
Its derivative $F'(\zeta)=\zeta-x_y+\sum_{j\ne y}[\zeta-x_j]_+$ is continuous and strictly increasing. If $k$ coordinates are below its root, then $(k+1)\zeta=x_y+\sum_{j\le k}x_{(j)}$. The minimum-prefix expression selects precisely such a consistent root, including ties, and strict convexity gives uniqueness. This formula does not assume $x_y=0$ when used for the full-KL gradient.
\paragraph{Nearest correction of the complete direction.}
Let $w\in\R^K$ be a candidate complete direction. The update $v$ in Eq.~\eqref{eq:core} is the unique solution of
\[
 \underset{w\in\R^K:\;w-h\in\Kc}{\operatorname{minimize}}\quad
 \frac12\|w-(h+u/2)\|^2.
\]
Indeed, the constraint preserves every margin gain of the corresponding label step, and translation by $h$ followed by the positive homogeneity of projection onto the cone gives $v=h+\Pi_{\Kc}(u)/2$.

\paragraph{Full task ordering.}
From $h=(1-\rho)(-1,r)$ we get $h_j-h_y=(1-\rho)(1+r_j)\ge0$ for every $j\ne y$, so $h\in\Kc$. Since $d\in\Kc$ by construction and $v-h=d/2\in\Kc$, for every $\eta\ge0$,
\[
 M_j(z-\eta v)\ge M_j(z-\eta h)\ge M_j(z).
\]
The cross-entropy ordering in Eq.~\eqref{eq:marginorder} follows from $\CE(z,y)=\log(1+\sum_{j\ne y}e^{-M_j(z)})$.

\subsection{Clipped teacher and the tight compensation bounds}\label{app:clipped-ce}
For the compensation in Eq.~\eqref{eq:clipped-core}, let $\TV(q,r)=\frac12\|q-r\|_1$, where $\|\cdot\|_1$ sums absolute coordinates. The tight size bounds are $\TV(q,r)/(K-1)\le\ell\le\TV(q,r)/2$: compensation is controlled by the mismatch between the two conditional distributions.
\paragraph{Proof.}
From the threshold form of the projection, Eq.~\eqref{eq:closedproj}, $d_y=-\ell$ and $d_j=\max(r_j-q_j,-\ell)$, and from $\one^\top d=0$ (Appendix~\ref{app:retentionproof}),
\[
 \ell=\sum_{j\ne y}\bigl[q_j-r_j-\ell\bigr]_+,\qquad
 d_j=r_j-\min(q_j,\ r_j+\ell).
\]
The function $g(a)=a-\sum_j[q_j-r_j-a]_+$ is continuous and strictly increasing. If $q\ne r$ then $g(0)=-\TV(q,r)<0$ and $g(\TV(q,r))>0$; if $q=r$ the unique root is zero. Hence $\ell$ is unique. Summing the clipped mass gives $\sum_j\widetilde q_j=1-\ell$.

Let $T=\TV(q,r)$ and let $k\le K-2$ be the number of coordinates with positive deviation ($q_j>r_j$). The root equation gives
\[
 T=\ell+\sum_{q_j>r_j}\min(q_j-r_j,\ \ell)\ \le\ (k+1)\ell\ \le\ (K-1)\ell .
\]
If $T>0$, the set $A=\{j:q_j-r_j>\ell\}$ is nonempty; with $m=|A|$,
\[
 T\ \ge\ \sum_{j\in A}(q_j-r_j)=(m+1)\ell\ \ge\ 2\ell .
\]
This proves the two compensation bounds. One positive deviation $a$ shared as negative deviation by the other coordinates gives $\ell=T/2$; $K-2$ equal positive deviations with the last coordinate carrying all the negative deviation gives $\ell=T/(K-1)$. Sufficiently small perturbations of the uniform distribution make both examples strictly positive, so both bounds are tight, and $\ell<1/2$.

Finally $d_j-d_y=[r_j-q_j+\ell]_+$. Substituting this into the margin change relative to the CE endpoint gives the extra-margin formula in Section~\ref{sec:clipped}.
\subsection{Proof of Theorem~\ref{thm:retention}}\label{app:retentionproof}
Let $n=K-1$. The Moreau decomposition gives $u=d+e$ with $e\in\Kc^\circ$, $d^\top e=0$, and $\norm u^2=\norm d^2+\norm e^2$. Vectors of the polar cone $\Kc^\circ$ have the form $(e)_y=\Lambda$, $(e)_j=-\lambda_j$, $\lambda_j\ge0$, $\Lambda=\sum_{j\ne y}\lambda_j$. Put $Q=\sum_j\lambda_j^2\le\Lambda^2$, so $\norm e^2=Q+\Lambda^2$. From $d^\top e=0$ we have $u^\top e=\norm e^2$; since $u_y=0$ and $\sum_{j\ne y}u_j=0$, the $(\lambda_j)$ may be centered inside the inner product:
\begin{align}
 \norm e^4=(u^\top e)^2=\Bigl(\sum_{j\ne y}u_j\bigl(\lambda_j-\tfrac{\Lambda}{n}\bigr)\Bigr)^2
 \le\norm u^2\Bigl(Q-\frac{\Lambda^2}{n}\Bigr)
 \le\frac{n-1}{2n}\norm u^2\,(Q+\Lambda^2),
\end{align}
where the last inequality is equivalent to $Q\le\Lambda^2$. If $e\ne0$, dividing by $\norm e^2$ gives $\norm e^2\le\frac{n-1}{2n}\norm u^2$ and hence $\norm d^2\ge\frac{n+1}{2n}\norm u^2=\gamma_K\norm u^2$; the case $e=0$ is trivial. Finally $u^\top d=(d+e)^\top d=\norm d^2$, which gives Eq.~\eqref{eq:retention}.

\paragraph{Pure conditional component.}
The polar representation gives $\one^\top e=0$, and with $\one^\top u=0$ we get $\one^\top d=0$. Let $\ell=-d_y$; then $\sum_{j\ne y}d_j=\ell$, and $w_{\mathrm{cond}}=d-\ell a_0$ satisfies $(w_{\mathrm{cond}})_y=0$ and $\one^\top w_{\mathrm{cond}}=0$, i.e.\ $w_{\mathrm{cond}}\in\Ls$; moreover $a_0^\top d=\ell(1+1/n)=\ell\norm{a_0}^2$ shows $w_{\mathrm{cond}}\perp a_0$. Since $u\perp a_0$, $u^\top w_{\mathrm{cond}}=u^\top d=\norm d^2$. For $u\ne0$, Cauchy--Schwarz gives
\[
 \norm{w_{\mathrm{cond}}}^2\ge\frac{(u^\top w_{\mathrm{cond}})^2}{\norm u^2}=\frac{\norm d^4}{\norm u^2}\ge\gamma_K\norm d^2\ge\gamma_K^2\norm u^2 .
\]

\paragraph{Tightness.}
Take $a>0$, $u_y=0$, one non-target coordinate $u_{j_*}=-a$, and the remaining $n-1$ coordinates equal to $a/(n-1)$. By Eq.~\eqref{eq:closedproj}, $\zeta=-a/2$, and the projection is $d_y=d_{j_*}=-a/2$, $d_j=a/(n-1)$ for $j\ne y,j_*$. A direct computation gives $\norm d^2/\norm u^2=(n+1)/(2n)=\gamma_K$ and $w_{\mathrm{cond}}=\gamma_Ku$, so both bounds are attained simultaneously; for sufficiently small $a$ this $u$ is the difference of two strictly positive conditional distributions.

\subsection{The alignment part of Theorem~\ref{thm:retention}}\label{app:alignmentproof}
Keep $u=d+e$ and the polar representation. Since $u_y=0$, $(e)_y=-d_y=\ell=\sum_j\lambda_j$, so $\norm e^2=\ell^2+\sum_j\lambda_j^2\le2\ell^2$. Write $\gamma=\gamma_K$. From $w_{\mathrm{cond}}\perp a_0$, $u\perp a_0$, and $\norm{a_0}^2=(n+1)/n=2\gamma$,
\[
 \norm d^2=\norm{w_{\mathrm{cond}}}^2+2\gamma\ell^2,\qquad u^\top w_{\mathrm{cond}}=\norm d^2 .
\]
Therefore
\begin{align}
 \gamma\norm u^2=\gamma\norm d^2+\gamma\norm e^2
 &\le\gamma\norm d^2+2\gamma\ell^2\notag\\
 &=\gamma\norm d^2+\norm d^2-\norm{w_{\mathrm{cond}}}^2
 =(1+\gamma)\,u^\top w_{\mathrm{cond}}-\norm{w_{\mathrm{cond}}}^2,
\end{align}
Completing the square in this inequality gives the equivalent form
\begin{equation}\label{eq:relation-ball}
 \Bigl\|w_{\mathrm{cond}}-\frac{1+\gamma}{2}u\Bigr\|^2\le\frac{(1-\gamma)^2}{4}\norm u^2 .
\end{equation}
For $u\ne0$ we have $w_{\mathrm{cond}}\ne0$, and $(1+\gamma)u^\top w_{\mathrm{cond}}\ge\gamma\norm u^2+\norm{w_{\mathrm{cond}}}^2\ge2\sqrt\gamma\norm u\norm{w_{\mathrm{cond}}}$ gives Eq.~\eqref{eq:relation-angle}. Since $\gamma\in(1/2,3/4]$ and $2\sqrt\gamma/(1+\gamma)$ is increasing on $(0,1]$, the uniform lower bound is $2\sqrt2/3$.

\subsection{The geometric example in Figure~\ref{fig:projection-geometry}}\label{app:geometry-example}
The illustration uses $K=4$, with the correct class first, $r=(0.20,0.30,0.50)$ and $q=(0.30,0.32,0.38)$. These give
\[
 u=(0,-0.10,-0.02,0.12),\qquad d=(-0.05,-0.05,-0.02,0.12),\qquad\ell=0.05.
\]
With $a_0=(-1,1/3,1/3,1/3)$, the pure conditional component is
\[
 w_{\rm cond}=(0,-1/15,-11/300,31/300).
\]
It is orthogonal to $a_0$ and forms an angle of approximately $11.54^\circ$ with $u$. Panel (a) uses an oblique view to display both orthogonal components; panel (b) shows the angle within the conditional plane. This is a concrete illustration of the decomposition and alignment in Theorem~\ref{thm:retention}.

\subsection{A local loss representation of the clipped teacher}\label{app:local-surrogate}
The clipped direction in Eq.~\eqref{eq:clipped-core} has an exact local loss interpretation. Fix a reference output $z_0$, compute $\ell$ and $\widetilde q$ there, and hold both fixed. Since $\sum_j\widetilde q_j=1-\ell>0$, the vector $\bar q=\widetilde q/(1-\ell)$ is a positive conditional distribution. Define
\begin{equation}\label{eq:local-clipped-loss}
 \Lambda_{z_0}(z)=(1-\ell)\KL(\bar q\Vert r(z))-\ell\logit\rho(z).
\end{equation}
Then
\begin{equation}\label{eq:local-clipped-gradient}
 \left.\nabla_z\Lambda_{z_0}(z)\right|_{z=z_0}=d(z_0).
\end{equation}
The two terms learn the normalized clipped teacher and increase correct-class log-odds, respectively, with weights determined by the retained and clipped probability masses.

\paragraph{Proof.}
Using $\logit\rho(z)=z_y-\log\sum_{j\ne y}e^{z_j}$ and differentiating with the reference quantities frozen gives
\[
 \partial_{z_y}\Lambda_{z_0}=-\ell,\qquad
 \partial_{z_j}\Lambda_{z_0}=(1-\ell)(r_j(z)-\bar q_j)+\ell r_j(z)
 =r_j(z)-\widetilde q_j.
\]
At $z=z_0$ these are precisely the coordinates of $d(z_0)$ in Eq.~\eqref{eq:clipped-core}.

\subsection{The exact signal removed and the size of compensation}\label{app:exact-retained}
The retention bound can be refined using the coordinates actually clipped. Let $\lambda=(\lambda_j)_{j\ne y}$ collect the clipped masses, defined by
\[
 \lambda_j=[q_j-r_j-\ell]_+,\qquad e=u-d=(\ell,-\lambda),\qquad
 \sum_{j\ne y}\lambda_j=\ell.
\]
Moreau orthogonality gives the exact amount removed from the conditional first-order signal:
\begin{equation}\label{eq:exact-removed}
 \|u\|^2-u^\top d=\|u\|^2-\|d\|^2
 =\ell^2+\sum_{j\ne y}\lambda_j^2.
\end{equation}
For $u\ne0$, let $m=|\{j:\lambda_j>0\}|$. Then
\begin{equation}\label{eq:active-removed}
 \ell^2(1+1/m)\le\|u\|^2-\|d\|^2\le2\ell^2.
\end{equation}
The lower bound follows from Cauchy--Schwarz applied to the $m$ positive coordinates of $\lambda$; the upper bound follows from their nonnegativity and sum $\ell$.

The extra task-side change is also controlled by the current conditional mismatch. Write $T=\TV(q,r)=\frac12\|q-r\|_1$ and let $\Delta M_j^{\rm extra}$ be the TPKD margin minus the corresponding CE-step margin. Because $r_j-q_j\le T$ and $\ell\le T/2$, Eq.~\eqref{eq:clipped-core} gives, for every $\eta\ge0$,
\begin{equation}\label{eq:extra-margin-bound}
 0\le\Delta M_j^{\rm extra}=\frac\eta2[r_j-q_j+\ell]_+\le\frac{3\eta T}{4}.
\end{equation}
Let $z'=z-\eta h$ and $\rho'=\softmax(z')_y$. Convexity of CE and the margin ordering imply
\begin{align}\label{eq:extra-ce-bound}
 0\le\CE(z')-\CE(z'-\eta d/2)
 &\le\frac\eta2 h(z')^\top d\notag\\
 &=\frac\eta2(1-\rho')\sum_{j\ne y}r_j(z')(d_j-d_y)\notag\\
 &\le\frac{3\eta}{4}(1-\rho')T
 \le\frac{3\eta}{4}(1-\rho)T.
\end{align}
Thus, at a fixed output step size, the additional confidence effect vanishes as the conditional mismatch vanishes.

\subsection{Optimal conditional progress under two local budgets}\label{app:local-optimality}
The nearest-safe projection also maximizes conditional progress under two complementary comparisons.

\paragraph{Local quadratic model.}
Let $L>0$ be a smoothness bound for $\Shape$ and let $x$ be an output displacement, so that $z^+=z-x$. Smoothness gives
\[
 \Shape(z)-\Shape(z-x)\ge Q_u(x),\qquad
 Q_u(x)=u^\top x-\frac L2\|x\|^2.
\]
Completing the square yields
\[
 Q_u(x)=\frac{\|u\|^2}{2L}-\frac L2\|x-u/L\|^2.
\]
The unconstrained maximizer is $u/L$; over $x\in\Kc$ it is $d/L$, by positive homogeneity of cone projection. Consequently,
\begin{equation}\label{eq:quadratic-optima}
 \max_{x\in\Kc}Q_u(x)=\frac{\|d\|^2}{2L},\qquad
 \max_xQ_u(x)=\frac{\|u\|^2}{2L}.
\end{equation}
For $u\ne0$, the ratio of these optimal guaranteed decreases is $\|d\|^2/\|u\|^2\ge\gamma_K$, with equality in the tightness construction of Appendix~\ref{app:retentionproof}.

\paragraph{Fixed displacement length.}
For $u\ne0$ and a budget $S>0$,
\begin{equation}\label{eq:fixed-budget-optimum}
 \max_{x\in\Kc,\,\|x\|\le S}u^\top x=S\|d\|,
 \qquad x^*=S\frac d{\|d\|}.
\end{equation}
To prove this, use $u=d+e$, where $e\in\Kc^\circ$: for any feasible $x$,
\[
 u^\top x=d^\top x+e^\top x\le\|d\|\|x\|\le S\|d\|.
\]
The stated $x^*$ attains equality. Without the cone constraint the optimum is $S\|u\|$, so the safe-to-unconstrained ratio is $\|d\|/\|u\|\ge\sqrt{\gamma_K}$, again tight. This is the comparison at equal displacement length; Eq.~\eqref{eq:retention} compares the original and projected directions at equal step size.

\section{Proofs supporting joint fitting and the CE comparison}\label{app:dynamics}
We prove Theorem~\ref{thm:convergence}, including its residual rate, and the uniform-limit comparison stated in Section~\ref{sec:dynamics}.
\subsection{Complete-update alignment in Theorem~\ref{thm:convergence}}\label{app:fullalignmentproof}
Write $u=d+e$ for the Moreau decomposition. Then $d^\top e=0$, $e\in\Kc^\circ$, and Theorem~\ref{thm:retention} implies
\[
 \norm e^2\le\frac{1-\gamma_K}{\gamma_K}\norm d^2.
\]
We have $h\in\Kc$, so $h^\top e\le0$. Also, using the CE form $h=(1-\rho)(-1,r)$,
\[
 h^\top d=(1-\rho)\sum_{j\ne y}r_j(d_j-d_y)\ge0.
\]
Set $g_\Phi=4h+u$ and abbreviate $\gamma=\gamma_K$. Expanding gives
\begin{align}\label{eq:alignment-expansion}
 g_\Phi^\top v-\norm v^2-\frac\gamma4\norm{g_\Phi}^2
 &=(3-4\gamma)\norm h^2+2(1-\gamma)h^\top d+(1-2\gamma)h^\top e\notag\\
 &\quad+\frac14\left[(1-\gamma)\norm d^2-\gamma\norm e^2\right]\ge0.
\end{align}
Every term is nonnegative because $1/2<\gamma\le3/4$. The arithmetic--geometric mean inequality gives
\[
 g_\Phi^\top v\ge\norm v^2+\frac\gamma4\norm{g_\Phi}^2
 \ge\sqrt\gamma\norm{g_\Phi}\norm v,
\]
which proves Eq.~\eqref{eq:full-alignment}.

\paragraph{Sharpness.}
Fix any strictly positive pair $r,q$ realizing the equality construction in Appendix~\ref{app:retentionproof}; for example, perturb the uniform conditional distribution by a sufficiently small one-negative-coordinate, equal-positive-remainder difference. Keep this $u=(0,r-q)$ fixed and let the student correct-class probability $\rho$ tend to one. Then $h\to0$, $g_\Phi\to u$ and $v\to d/2$, whence
\[
 \cos\angle(g_\Phi,v)\ \longrightarrow\
 \frac{u^\top d}{\norm u\norm d}
 =\frac{\norm d}{\norm u}=\sqrt\gamma.
\]
Thus no larger uniform angle constant holds for strictly positive probabilities. This construction also makes the gap in Eq.~\eqref{eq:alignment-expansion} tend to zero.

\paragraph{Batch form.}
Let $\operatorname{col}$ denote vertical stacking. For $W_\Phi=\operatorname{col}(g_{\Phi,i})/\sqrt N$ and $V=\operatorname{col}(v_i)/\sqrt N$, summing the per-example inequality yields
\begin{equation}\label{eq:sector}
 W_\Phi^\top V\ge\norm V^2+\frac\gamma4\norm{W_\Phi}^2
 \ge\sqrt\gamma\norm{W_\Phi}\norm V.
\end{equation}

\subsection{Descent, convergence and the residual rate}\label{app:dynamicsproof}
The logit Hessians of the cross-entropy and conditional KL are softmax covariance matrices, with spectral norm at most $1/2$. Therefore $\Phi=4\CE+\Shape$ is $5/2$-smooth. For the specified output step,
\begin{align}
 \Phi(z-\eta v)&\le\Phi(z)-\eta g_\Phi^\top v+\frac54\eta^2\norm v^2\notag\\
 &\le\Phi(z)-\frac{\eta\gamma_K}{4}\norm{g_\Phi}^2
 -\eta\left(1-\frac54\eta\right)\norm v^2.
\end{align}
This proves Eq.~\eqref{eq:descentcertificate}. When $0<\eta\le4/5$, the last term is nonpositive. For any integer $T\ge1$, summing over $k=0,\ldots,T-1$ and using $\Phi\ge0$ gives
\begin{equation}\label{eq:residual-rate}
 \sum_{k<T}\norm{g_{\Phi,k}}^2\le\frac{4\Phi(z_0)}{\eta\gamma_K}.
\end{equation}
Thus $g_{\Phi,k}\to0$. Since $u_{k,y}=0$ and $(g_{\Phi,k})_y=-4(1-\rho_k)$, we obtain $\rho_k\to1$. The identity $h_k=(1-\rho_k)(-1,r_k)$ then gives $h_k\to0$, so $u_k=g_{\Phi,k}-4h_k\to0$ and $r_k\to q$. Finally $\norm{d_k}\le\norm{u_k}$ implies $v_k\to0$.

\paragraph{Joint and component residuals.}
Dividing Eq.~\eqref{eq:residual-rate} by $T$ gives
\begin{equation}\label{eq:average-joint-residual}
 \frac1T\sum_{k<T}\|g_{\Phi,k}\|^2\le\frac{4\Phi(z_0)}{\eta\gamma_K T}.
\end{equation}
The two learning signals inherit quantitative bounds. Since $h=(1-\rho)(-1,r)$ and $(g_\Phi)_y=-4(1-\rho)$,
\[
 \|h\|^2=(1-\rho)^2(1+\|r\|^2)\le2(1-\rho)^2\le\frac18\|g_\Phi\|^2.
\]
Also $u=g_\Phi-4h$ implies $\|u\|^2\le2\|g_\Phi\|^2+32\|h\|^2\le6\|g_\Phi\|^2$. Hence
\begin{equation}\label{eq:average-component-residual}
 \frac1T\sum_{k<T}\left(\|h_k\|^2+\frac1{12}\|u_k\|^2\right)
 \le\frac{5\Phi(z_0)}{2\eta\gamma_K T}.
\end{equation}
These bounds quantify the vanishing time-averaged squared gradient residuals of the joint, label and conditional signals.
\subsection{The CE conditional limit stated in Section~\ref{sec:dynamics}}\label{app:ceuniform}
Let $n=K-1$, $\one_n=(1,\ldots,1)\in\R^n$, and $\nu=\one_n/n$. From finite initial logits, CE gradient flow and the fixed-step rule with $0<\eta\le1/20$ have limits $p_y\to1$ and $r\to\nu$. This differs from the TPKD conditional limit only when $q\ne\nu$.
\paragraph{Proof.}
Let $z_{\neg y}\in\R^n$ contain the non-target logits, set $a=z_{\neg y}-(\one_n^\top z_{\neg y})\one_n/n$, and define on the zero-sum subspace
\[
 F(a)=\log\sum_je^{a_j}-\frac1n\sum_ja_j-\log n=\KL(\nu\Vert r),\qquad\nabla F=r-\nu .
\]
Let $t\ge0$ denote flow time, with dots denoting time derivatives. Along the cross-entropy gradient flow, $\dot z_y=1-\rho$ and $\dot z_j=-(1-\rho)r_j$, so
\[
 \dot a=-(1-\rho)\nabla F,\qquad\dot F=-(1-\rho)\norm{r-\nu}^2,
\]
which gives the conditional dynamics of CE. Let $s(t)=\int_0^t(1-\rho(\tau))\,d\tau$. If $s(\infty)<\infty$, every logit has finite total variation, all logits converge to finite values, and $1-\rho$ converges to a positive number, contradicting the finiteness of the integral; hence $s(t)\to\infty$. In the effective time $s$, $da/ds=-\nabla F$. The sublevel sets of $F$ on the zero-sum subspace are bounded and its Hessian $\operatorname{diag}(r)-rr^\top$, where $\operatorname{diag}(r)$ is the diagonal matrix with diagonal $r$, is positive definite on that subspace, so on the compact sublevel set containing the whole trajectory there is $\mu>0$ such that $F$ is $\mu$-strongly convex. The unique minimizer is $a=0$, hence $r\to\nu$. Moreover $z_y=z_y(0)+s(t)\to\infty$ while all non-target logits are nonincreasing, so $\rho\to1$.

In the discrete case let $s_k=\eta(1-\rho_k)$, so $a_{k+1}=a_k-s_k\nabla F(a_k)$. The gradient of $F$ is $\frac12$-Lipschitz, hence
\[
 F(a_{k+1})\le F(a_k)-s_k(1-s_k/4)\norm{\nabla F(a_k)}^2 .
\]
For $0<\eta\le1/20$ the sequence stays in the same compact sublevel set. If $\sum_ks_k<\infty$, the same finite-total-variation argument gives a positive limit of $1-\rho_k$, a contradiction; hence $\sum_ks_k=\infty$. By strong convexity on that compact set and $\norm{\nabla F}^2\ge2\mu F$,
\[
 F(a_{k+1})\le\bigl[1-2\mu s_k(1-s_k/4)\bigr]F(a_k),
\]
so $F(a_k)\to0$ and $r_k\to\nu$. At the same time $z_{k,y}=z_{0,y}+\sum_{j<k}s_j\to\infty$ and all non-target logits are nonincreasing, so $\rho_k\to1$.

\subsection{When a complete step lowers its own conditional error}\label{app:conditional-descent}
The effect of CE on the teacher's conditional objective has the exact form
\begin{equation}\label{eq:ce-shape-interaction}
 u^\top h=(1-\rho)(r-q)^\top r
 =\frac{1-\rho}{2}\left(\|r-q\|^2+\|r\|^2-\|q\|^2\right).
\end{equation}
This follows by expanding $\|r-q\|^2$. The projected conditional contribution adds $\|d\|^2/2$, so for the complete direction $v=h+d/2$,
\begin{equation}\label{eq:own-shape-direction}
 u^\top v=u^\top h+\tfrac12\|d\|^2
 \ge-\|h\|\|u\|+\frac{\gamma_K}{2}\|u\|^2.
\end{equation}
In particular, $\|u\|>0$ and $\|h\|/\|u\|<\gamma_K/2$ imply $c=u^\top v>0$. The conditional KL has logit Hessian $\operatorname{diag}(r)-rr^\top$ on the non-target coordinates, with spectral norm at most $1/2$. Thus
\[
 \Shape(z-\eta v)\le\Shape(z)-\eta c+\frac{\eta^2}{4}\|v\|^2.
\]
Consequently the sufficient conditions
\begin{equation}\label{eq:own-shape-condition}
 \|u\|>0,\qquad \frac{\|h\|}{\|u\|}<\frac{\gamma_K}{2},\qquad
 0<\eta<\frac{4u^\top v}{\|v\|^2}
\end{equation}
give $\Shape(z-\eta v)<\Shape(z)$. The complete update then reduces its own conditional error while retaining the label-step margin gains in Eq.~\eqref{eq:marginorder}.

\subsection{General positive conditional weights}\label{app:general-beta}
For positive coefficients $\beta,w$, consider the family
\[
 v_\beta=h+\beta d,\qquad \Phi_w=w\CE+\Shape.
\]
For every $\beta>0$, $v_\beta-h=\beta d\in\Kc$, so the label-step margin ordering holds for all $\eta\ge0$. Moreover, $v_\beta$ is the unique minimizer of $\frac12\|a-(h+\beta u)\|^2$ over candidate directions $a\in\R^K$ with $a-h\in\Kc$.

A sufficient joint-learning condition is
\begin{equation}\label{eq:general-coefficient-condition}
 4w\beta\gamma_K>1.
\end{equation}
To make the resulting constants explicit, set
\[
 c_{w,\beta}=\frac{w+\beta\gamma_K-
 \sqrt{(w-\beta\gamma_K)^2+1}}{2}>0,\qquad
 C_\beta=\max\{1,\beta^2\}.
\]
For a fixed positive conditional target and finite initial logits, every constant step
\begin{equation}\label{eq:general-coefficient-step}
 0<\eta\le\frac{c_{w,\beta}}{(w+1)C_\beta}
\end{equation}
satisfies
\begin{equation}\label{eq:general-coefficient-descent}
 \Phi_w(z-\eta v_\beta)\le\Phi_w(z)
 -\frac{\eta c_{w,\beta}}2\bigl(\|h\|^2+\|u\|^2\bigr).
\end{equation}
It follows that
\[
 \frac1T\sum_{k<T}\bigl(\|h_k\|^2+\|u_k\|^2\bigr)
 \le\frac{2\Phi_w(z_0)}{\eta c_{w,\beta}T},
 \qquad p_{k,y}\to1,\quad r_k\to q,\quad v_{\beta,k}\to0.
\]

\paragraph{Proof.}
Write $H=\|h\|$ and $S=\|u\|$. The identities $h^\top d\ge0$ and $u^\top d\ge\gamma_K\|u\|^2$ give
\[
 \nabla\Phi_w^\top v_\beta
 =w\|h\|^2+w\beta h^\top d+u^\top h+\beta u^\top d
 \ge wH^2-HS+\beta\gamma_K S^2.
\]
The matrix of this quadratic form is
$\left(\begin{smallmatrix}w&-1/2\\-1/2&\beta\gamma_K\end{smallmatrix}\right)$.
It is positive definite under Eq.~\eqref{eq:general-coefficient-condition}, with smallest eigenvalue $c_{w,\beta}$, so the last expression is at least $c_{w,\beta}(H^2+S^2)$. The potential is $(w+1)/2$-smooth and
\[
 \|v_\beta\|^2\le2H^2+2\beta^2S^2\le2C_\beta(H^2+S^2).
\]
The descent lemma therefore gives
\[
 \Phi_w(z-\eta v_\beta)\le\Phi_w(z)
 -\eta\left[c_{w,\beta}-\frac{(w+1)C_\beta\eta}{2}\right](H^2+S^2),
\]
which implies Eq.~\eqref{eq:general-coefficient-descent}. Summation and $\Phi_w\ge0$ show that $h_k,u_k\to0$. The identity $h_{k,y}=p_{k,y}-1$ gives $p_{k,y}\to1$, while $u_k=(0,r_k-q)$ gives $r_k\to q$; finally $\|d_k\|\le\|u_k\|$ yields $v_{\beta,k}\to0$.

\section{Proofs supporting exact realization and parameter learning}\label{app:parameters}
This appendix proves Proposition~\ref{prop:exact}, Theorem~\ref{thm:parameter}, and the measured-angle test stated in Section~\ref{sec:realization}.

\subsection{Exact head realization}\label{app:exactproof}
For $N$ frozen features of dimension $D$, absorb the classifier bias into the last feature column. Let $X\in\R^{N\times D}$ be the feature matrix, $A\in\R^{D\times K}$ the classifier weights, and $Z=XA$ the batch logits. Let $V_{\rm row}$ have rows $v_i^\top$, $X^\dagger$ denote the pseudoinverse, $I_N$ the $N\times N$ identity, and $\Delta A$ the head-weight displacement. The exact construction in Proposition~\ref{prop:exact} is
\begin{equation}\label{eq:exact}
 \Delta A=-\eta X^\dagger V_{\rm row},\qquad
 X(A+\Delta A)=Z-\eta V_{\rm row}.
\end{equation}
 Full row rank gives $XX^\dagger=I_N$, so $X(A-\eta X^\dagger V_{\rm row})=Z-\eta V_{\rm row}$. Let $H\in\R^{D\times K}$ satisfy $XH=0$. Every other feasible displacement is $-\eta X^\dagger V_{\rm row}+H$. The columns of $X^\dagger V_{\rm row}$ lie in the column space of $X^\top$ and are orthogonal to the columns of $H$; writing $\|\cdot\|_F$ for the Frobenius norm gives
\[
 \|-\eta X^\dagger V_{\rm row}+H\|_F^2
 =\eta^2\|X^\dagger V_{\rm row}\|_F^2+\|H\|_F^2.
\]
This proves the unique minimum-norm claim and the exact inheritance of the same step's output orderings.

\paragraph{Repeated realization on fixed features.}
Let the same full-row-rank $X$ be used at every iteration, with fixed positive conditional targets for its rows. For $A_{k+1}=A_k-\eta X^\dagger V_{{\rm row},k}$,
\[
 Z_{k+1}=X A_{k+1}=Z_k-\eta V_{{\rm row},k}.
\]
Thus every row follows the output iteration of Theorem~\ref{thm:convergence}. With finite initial logits and $0<\eta\le4/5$, each row has $p_{i,k,y_i}\to1$ and $r_{i,k}\to q_i$.

\paragraph{Computational cost.}
For $N\le D$, forming and factorizing $XX^\top$ and applying $X^\dagger V_{\rm row}$ costs $O(N^2D+N^3+NDK)$. With fixed $X$, its pseudoinverse can be reused, leaving $O(NDK)$ work per head displacement.

\paragraph{Parameter steps and batch objectives.}
Let $g_{\rm T}=N^{-1}\sum_iJ_i^\top v_i$ and $g_{\rm C}=N^{-1}\sum_iJ_i^\top(h_i+\ell_i a_{0,i}/2)$ be the respective parameter gradients of TPKD and compensation only. Define their conditional increment $\delta g=g_{\rm T}-g_{\rm C}$, the batch-average conditional loss $\overline L_{\rm cond}(\theta)=N^{-1}\sum_i\Shape(z_i(\theta))$, and the batch-average joint potential $\Psi(\theta)=N^{-1}\sum_i\Phi(z_i(\theta))$. Under the respective conditions of Theorem~\ref{thm:parameter}, for sufficiently small $\eta>0$ its conclusions are
\begin{equation}\label{eq:parameter-conclusions}
 \begin{aligned}
 \textup{(i)}\quad &\delta g\ne0,\qquad \overline L_{\rm cond}(\theta-\eta g_{\rm T})<\overline L_{\rm cond}(\theta-\eta g_{\rm C}),\\
 \textup{(ii)}\quad &\Psi(\theta-\eta g_{\rm T})<\Psi(\theta).
 \end{aligned}
\end{equation}

\subsection{The angle--spectrum argument}\label{app:sectorproof}
\paragraph{Batch signals and restricted spectra.}\label{app:parameter-signals}
Here $\operatorname{col}$ denotes vertical concatenation. For the batch Jacobian $J=\operatorname{col}(J_i)/\sqrt N$, define $G=JJ^\top$ and the normalized signal stacks
\[
 (U,R)=\frac1{\sqrt N}\bigl(\operatorname{col}(u_i),\operatorname{col}(w_{{\rm cond},i})\bigr),
 \qquad
 (W,V)=\frac1{\sqrt N}\bigl(\operatorname{col}(g_{\Phi,i}),\operatorname{col}(v_i)\bigr).
\]
The first pair describes conditional learning and the second joint descent. Equation~\eqref{eq:surrogate} yields
\[
 \delta g=J^\top R/2,\qquad \nabla_\theta\overline L_{\rm cond}=J^\top U,
 \qquad \nabla_\theta\Psi=J^\top W,\qquad g_{\rm T}=J^\top V.
\]
For either pair let $Q$ have orthonormal columns spanning its subspace. The smallest and largest eigenvalues of $Q^\top GQ$ are the squared-stretch extrema $m,M$ for $J^\top$. Use $(m_c,M_c)$ for $\operatorname{span}\{U,R\}$ and $(m_\Phi,M_\Phi)$ for $\operatorname{span}\{W,V\}$. The batch map in Section~\ref{sec:realization} is $J^\top/\sqrt N$; its squared stretches are $m/N,M/N$, with the same ratio. Consequently $\chi_c=M_c/m_c$ and $\chi_\Phi=M_\Phi/m_\Phi$ when their minima are positive, and the ratios are infinite otherwise. The limits used in Theorem~\ref{thm:parameter} are
\begin{equation}\label{eq:conditional-kernel-condition}
 \kappa_c=\left(\frac{1+\sqrt{\gamma_K}}{1-\sqrt{\gamma_K}}\right)^2,
 \qquad \kappa_\Phi=\frac{1+\sqrt{\gamma_K}}{1-\sqrt{\gamma_K}}.
\end{equation}
With $K=100$, these give $\kappa_c\approx34.957650$ and $\kappa_\Phi\approx5.912499$.

Let $a,b\ne0$ have cosine $c$, and suppose the compression of $G=JJ^\top$ to their span has spectrum in $[m,M]$, $m>0$. For $-1<c<1$, the unit bisectors
\[
 e_\pm=\frac{a/\|a\|\pm b/\|b\|}{\sqrt{2(1\pm c)}}
\]
are orthonormal. Symmetry of $G$ cancels the cross terms and gives
\begin{equation}\label{eq:actual-angle-bound}
 \frac{a^\top Gb}{\|a\|\|b\|}
 =\tfrac12\bigl[(1+c)e_+^\top Ge_+-(1-c)e_-^\top Ge_-\bigr]
 \ge\tfrac12[(1+c)m-(1-c)M].
\end{equation}
Thus a positive measured cosine certifies $a^\top Gb>0$ whenever $M/m<(1+c)/(1-c)$. For $c=1$, $a,b$ are positively collinear and the lower bound is $m\|a\|\|b\|>0$. This proves the measured-angle refinement described in Section~\ref{sec:realization}. Using a universal lower bound on $c$ gives the class-dependent tests instead.

\subsection{Conditional increment and finite-step advantage}\label{app:backproprelation}
Assume $U\ne0$ and let the conditional signal subspace have spectral bounds $0<m_c\le M_c$. Let $\delta g=g_{\rm T}-g_{\rm C}=J^\top R/2$. Averaging the per-example squared-norm bounds gives $\|R\|^2\ge\gamma_K^2\|U\|^2$. To transfer the angle bound to the stacked signals, average the stronger inequality proved in Appendix~\ref{app:alignmentproof}:
\begin{equation}\label{eq:stacked-conditional-alignment}
 (1+\gamma_K)U^\top R\ge\gamma_K\|U\|^2+\|R\|^2
 \ge2\sqrt{\gamma_K}\|U\|\|R\|.
\end{equation}
This yields $R\ne0$ and $\cos\angle(U,R)\ge c_K$. Therefore
\begin{equation}\label{eq:parameter-shape-gain}
 \|\delta g\|^2=\tfrac14 R^\top GR\ge\tfrac{m_c}{4}\|R\|^2
 \ge\tfrac{m_c\gamma_K^2}{4}\|U\|^2>0.
\end{equation}
Moreover, with $\beta_c=(1+c_K)m_c-(1-c_K)M_c$, Eq.~\eqref{eq:actual-angle-bound} yields
\[
 t_c:=\ip{\nabla_\theta\overline L_{\rm cond}}{\delta g}
 =\tfrac12 U^\top GR\ge\tfrac{\beta_c}{4}\|U\|\|R\|.
\]
The threshold $M_c/m_c<(1+c_K)/(1-c_K)$ makes $t_c>0$. Since $c_K=2\sqrt{\gamma_K}/(1+\gamma_K)$, this threshold is exactly $\kappa_c$ in Eq.~\eqref{eq:conditional-kernel-condition}. At a differentiable parameter state,
\[
 \overline L_{\rm cond}(\theta-\eta g_{\rm T})-\overline L_{\rm cond}(\theta-\eta g_{\rm C})
 =-\eta t_c+o(\eta)<0
\]
for sufficiently small positive $\eta$, proving the conditional conclusion in Eq.~\eqref{eq:parameter-conclusions}. More explicitly, if the loss gradient is $L_c$-Lipschitz on both step segments, the difference is at most $-\eta t_c+L_c\eta^2(\|g_{\rm T}\|^2+\|g_{\rm C}\|^2)/2$; choosing $\eta\le t_c/[L_c(\|g_{\rm T}\|^2+\|g_{\rm C}\|^2)]$ gives an upper bound of $-\eta t_c/2$.

\subsection{Joint-potential descent}\label{app:jointproof}
Assume $W,V\ne0$ and let the joint signal subspace have spectral bounds $0<m_\Phi\le M_\Phi$. The batch form of Eq.~\eqref{eq:full-alignment} gives $\cos\angle(W,V)\ge s_K=\sqrt{\gamma_K}$. Applying Eq.~\eqref{eq:actual-angle-bound} on $\operatorname{span}\{W,V\}$ gives
\[
 t_\Phi:=\ip{\nabla_\theta\Psi}{g_{\rm T}}=W^\top GV
 \ge\tfrac12[(1+s_K)m_\Phi-(1-s_K)M_\Phi]\|W\|\|V\|.
\]
Thus $M_\Phi/m_\Phi<\kappa_\Phi$ makes $t_\Phi>0$, and $\Psi(\theta-\eta g_{\rm T})=\Psi(\theta)-\eta t_\Phi+o(\eta)<\Psi(\theta)$ for sufficiently small positive $\eta$. This proves Eq.~\eqref{eq:parameter-conclusions}.

\paragraph{An explicit finite-step bound.}
Put
\[
 E=\frac1N\sum_i\left(\|h_i\|^2+\frac1{12}\|u_i\|^2\right),\qquad
 \mu_\Phi=\frac12[(1+\sqrt{\gamma_K})m_\Phi-(1-\sqrt{\gamma_K})M_\Phi].
\]
Under $M_\Phi/m_\Phi<\kappa_\Phi$, $\mu_\Phi>0$. The output inequalities give
\[
 (4h+u)^\top v\ge4\|h\|^2-\|h\|\|u\|+\tfrac14\|u\|^2
 \ge\|h\|^2+\tfrac1{12}\|u\|^2,
\]
where $\|h\|\|u\|\le3\|h\|^2+\|u\|^2/12$ suffices. Hence $\|W\|\|V\|\ge W^\top V\ge E$, while $\|V\|^2\le8E$. It follows that
\[
 t_\Phi\ge\mu_\Phi E,\qquad
 \|g_{\rm T}\|^2=V^\top GV\le M_\Phi\|V\|^2\le8M_\Phi E.
\]
If $\nabla_\theta\Psi$ is $L_\Psi$-Lipschitz on the current step segment, with $L_\Psi>0$, then
\[
 \Psi(\theta-\eta g_{\rm T})\le\Psi(\theta)-\eta\mu_\Phi E
 +4L_\Psi M_\Phi\eta^2E.
\]
Consequently,
\begin{equation}\label{eq:explicit-parameter-step}
 0<\eta\le\frac{\mu_\Phi}{8M_\Phi L_\Psi}
 \quad\Longrightarrow\quad
 \Psi(\theta-\eta g_{\rm T})\le\Psi(\theta)-\frac{\eta\mu_\Phi}{2}E.
\end{equation}

\subsection{Computing the two-dimensional certificates}\label{app:certificate-computation}
For either output pair, orthonormalize its span to obtain $Q=[\xi_1,\xi_2]$, omitting a dependent column. Compute $b_i=J^\top \xi_i$ by vector--Jacobian products; the compressed matrix has entries $(Q^\top GQ)_{ij}=b_i^\top b_j$. Its smallest and largest eigenvalues are the required $m,M$. No full Jacobian or kernel matrix is formed. The direct conditional and joint inner products are $\langle J^\top U,J^\top R/2\rangle$ and $\langle J^\top W,J^\top V\rangle$, respectively. Each pair uses its own subspace, spectrum and measured angle.

\section{Experimental protocols}\label{app:experimental-details}
This appendix specifies the training and measurements reported in Section~\ref{sec:experiments}.
\subsection{Models, data, optimization and evaluation}\label{app:setup}
\paragraph{Vision.}
All visual experiments use CIFAR-100 with 50,000 training images and 10,000 test images, a VOLO-D2 teacher and a PiT-B student. The student combines a publicly pretrained backbone with a classifier fitted on features of the 50,000 training images by standardized logistic regression with regularization parameter $C=0.01$. All full-training methods use the same initial assets. Inputs use eight fixed augmented views: pad each 32-pixel image by four pixels, crop and horizontally flip, then bicubic resize to $224\times224$.

Full training uses 60 epochs, effective batch size 64 (microbatch 32), native automatic mixed precision (AMP), and SGD with momentum 0.9 and constant learning rate $10^{-4}$, without warmup or decay. Weight decay is $10^{-3}$, except for KL-Dist and XE-KL, whose recorded baseline configurations use $10^{-4}$. Every method is evaluated at epoch 60.

\paragraph{Text.}
CLINC150 has 150 in-domain classes. We combine its original 15,000 training examples and original 3,000 validation examples into one fixed 18,000-example training pool. The official 4,500-example in-domain test set remains separate. All methods use this same split and a fixed final-epoch evaluation.

The teacher is the existing five-epoch BERT-large. The student uses a publicly pretrained BERT-Mini backbone and a shared random classifier, initialized with truncated normal weights (standard deviation 0.02, limits $\pm0.04$) and zero bias. The head is trained from the first update. Inputs have maximum length 128 with fixed padding and an attention mask; the classifier receives the pretrained pooler output.

Training uses four epochs, 1,125 optimizer iterations, effective batch size 64 (microbatch 32), FP32 and disabled TF32. Google AdamWeightDecay uses moment-decay coefficients $(\beta_1,\beta_2)=(0.9,0.999)$ and numerical stabilizer $\epsilon=10^{-6}$, weight decay 0.01, no bias correction, and global gradient-norm clipping at 1.0. Bias and LayerNorm parameters are excluded from weight decay. With zero-indexed iteration $k$, the shared learning-rate schedule is
\[
 \eta_k=3\times10^{-4}\begin{cases}k/112,&k<112,\\1-k/1125,&112\le k<1125.\end{cases}
\]
Thus $3\times10^{-4}$ is the base rate of the common warmup and decay schedule.

\paragraph{Method-specific objectives and temperatures.}
The comparisons align data, teacher--student pair, initialization, training budget and evaluation within each domain. The loss used by each baseline retains its method-specific definition and prescribed scale. In particular, temperature is part of that definition, rather than a common training-budget parameter. The following choices are fixed in the reported runs.

KD and DKD use temperature $T=4$, matching the KD and DKD settings in the official DKD implementation \citep{zhao2022decoupled}. Its configuration declares \texttt{KD.TEMPERATURE=4} and \texttt{DKD.T=4.0}.\footnote{Official configuration: \url{https://github.com/megvii-research/mdistiller/blob/master/mdistiller/engine/cfg.py}.} For DKD, target and non-target weights are 1 and 8, respectively; the recorded method warmup is retained (two epochs in text). Standard KD uses distillation coefficient 1 in the shared training setup.

DHKD uses temperature 2 for its binary-KL objective, following its released training commands \citep{yang2025dhkd}; the official ImageNet command explicitly specifies \texttt{--BinaryKL\_T 2}.\footnote{Official DHKD training commands: \url{https://github.com/penghui-yang/DHKD}.} This is the temperature of DHKD's binary-KL loss, rather than a replacement of its objective by the KD loss.

The recorded KL-Dist, XE-KL, DP-U, DP-S and DTO-KD objectives use unit logit scale ($T=1$ where a temperature argument is present). TPKD also uses the original, unit-temperature conditional probabilities, with coefficient $1/2$. CE has no distillation temperature. Text DTO-KD retains its multi-layer feature-distillation implementation. These objective settings are unchanged across the three seeds; the five component controls use the TPKD settings and alter only the direction shown in Table~\ref{tab:ablation}.

\paragraph{Accuracy reporting.}
All full-training entries use seeds 42, 43 and 44. Within each domain, the comparisons share initial assets, training objects, data-order rules and the final-epoch evaluation. Tables~\ref{tab:application}--\ref{tab:ablation} report test accuracy in percent as the mean and sample standard deviation over $n=3$ seeds (denominator $n-1$). The final epoch is fixed before training; no validation, observation or best-test checkpoint is used for selection.

\paragraph{KL-Dist and DP-U on CLINC150.}
In the reported CLINC150 runs, DP-U selected the teacher target at a rate of 100\% in every epoch for all three seeds (42, 43 and 44), making its training objective identical to that of KL-Dist in these runs. Under the shared initialization, data order and optimization settings, we verified that the final logits were elementwise identical between the two methods for each seed. This explains their identical per-seed accuracies and the same mean and sample standard deviation of $93.38 \pm 0.19$\% in Table~\ref{tab:application}.

\subsection{Blocking and exact head updates}\label{app:exactsetup}
\paragraph{Blocking (Table~\ref{tab:states}).}
CE20, CE40 and CE60 are fixed seed-42 CE states under the visual protocol. Each uses 2,000 predetermined images with eight fixed views. The blocking criterion is $p_j\le t_j$ for every $j\ne y$. Conditional KL and $D_\infty$ are evaluated from log probabilities within the blocked set. Positive conditional error is checked at tolerance $10^{-12}$; all 16,000 image--view pairs enter the measurement.

\paragraph{Exact updates (Table~\ref{tab:exact}).}
Freeze the backbone features and absorb the head bias into the feature matrix. The head, direction calculation and SVD pseudoinverse use FP64. Each candidate starts from the same model and batch with output step $\eta=0.01$. The row-stacked direction specifies each example's output displacement, so it is not divided by batch size. The full-row-rank solve uses the pseudoinverse directly without ridge regularization.

The five directions are CE $h$, compensation only $h+\ell a_0/2$, unprojected conditional learning $h+u/2$, TPKD $h+d/2$, and CE plus full KL $h+(p-t)$. The last uses the original teacher probabilities with coefficient 1. Two states and 32 batches of 64 examples per state give 320 candidate steps, with two repeat checks. Let $z^+_{\rm CE}$ and $z^+_{\rm TPKD}$ denote the respective endpoint logits. The paired gain is $G_{\rm CE}=\Shape(z^+_{\rm CE})-\Shape(z^+_{\rm TPKD})$; table losses and changes are means over examples.

Retention is $u^\top d/\|u\|^2$ and the complete-update cosine is $\cos\angle(g_\Phi,v)$. We also verify the alignment residual $R_h=g_\Phi^\top v-\|v\|^2-\gamma_K\|g_\Phi\|^2/4$ and the finite-step descent bound in Eq.~\eqref{eq:descentcertificate}. A tolerance of $10^{-12}$ is used for inequalities and margin-order checks. The largest output execution error and CE-reference margin discrepancy are both $1.42\times10^{-14}$, supporting the numerical-precision statement in Section~\ref{sec:mechanism}.

\subsection{Backpropagation conditions and native paired updates}\label{app:certificates}
Table~\ref{tab:certificate} uses sixteen fixed batches at each of CE20 and CE60. TPKD uses $h+d/2$; compensation only uses $h+\ell a_0/2$, preserving the common confidence contribution while deleting the pure conditional component. Batch-average losses and the $1/\sqrt{64}$ normalization in Appendix~\ref{app:parameters} are used throughout.

Network and vector--Jacobian calculations use FP32, with FP64 geometric calculations and diagnostic microbatch size eight. The two restricted spectra are computed as in Appendix~\ref{app:certificate-computation}, with relative rank tolerance $10^{-10}$ and spectral tolerance $10^{-12}$. For 100 classes, the theoretical constants are $\gamma_{100}=100/198$, $\sqrt{\gamma_{100}}\approx0.7107$, $\kappa_c\approx34.96$ and $\kappa_\Phi\approx5.91$.

Each native candidate restores the same model, optimizer history, AMP scaler, buffers and random state, then executes the usual training iteration with microbatch size 32 and learning rate $10^{-4}$. Let $\theta^+_{\rm C}$ and $\theta^+_{\rm T}$ denote the resulting compensation-only and TPKD parameter endpoints. Their paired gain is
\[
 G_{\rm native}=\overline L_{\rm cond}(\theta^+_{\rm C})-\overline L_{\rm cond}(\theta^+_{\rm T}).
\]
FP32 paired replay evaluates both endpoint losses consistently. The geometric tests establish a useful conditional parameter direction; the native pairs measure its additional benefit under the training optimizer.

\subsection{The 128-iteration continuation}\label{app:window}
CE, compensation only and TPKD continue from the same CE60 state for 128 native-optimizer iterations at constant learning rate $10^{-4}$. They follow the same sequence of 8,192 update images, recomputing their directions from their own current student states. A fixed 2,048-image observation set is disjoint from these updates. Both sets are drawn from the original training pool, and observation is used only to measure conditional learning. We record conditional KL at iterations 0, 1, 8, 32, 64 and 128. Section~\ref{sec:mechanism} reports its starting value and TPKD's endpoint advantages.

\subsection{The five component controls}\label{app:ablation-setup}
All five rows in Table~\ref{tab:ablation} use the full-training protocols above. CE ($h$) and TPKD ($h+d/2$) use the corresponding main-table runs. Removing projection gives $h+u/2$. Removing the CE gradient gives $d/2$: it retains coefficient $1/2$, true-label indexing of the non-target classes, and the same safe cone. Safe full KL gives $h+\Pi_{\Kc}(p-t)/2$, using the full unit-temperature gradient and the generic projection in Eq.~\eqref{eq:closedproj}. Every control recomputes its direction at each iteration. The initialization, teacher, data order, optimization schedule and endpoint selection are unchanged.

\subsection{Implementation of the prescribed direction}\label{app:modules}
Keep the teacher fixed and detached. Compute $q=\softmax(z^T_{\neg y})$ and $r=\softmax(z_{\neg y})$ directly on non-target logits, form $u=(0,r-q)$, project to $d$, and set $v=h+d/2$. Detach $v$ in Eq.~\eqref{eq:surrogate}; its batch-averaged surrogate supplies the complete parameter data gradient, including the label contribution. Log-softmax values are used for conditional losses and compensation statistics, avoiding division by small $1-p_y$ values. The native optimizer then applies its usual update to the student parameters.

\end{document}